# Recreation of the Periodic Table with an Unsupervised Machine Learning Algorithm

**Minoru Kusaba[1,*], Chang Liu[2], Yukinori Koyama[3], Kiyoyuki Terakura[4], Ryo Yoshida[1,2,3,*]**

[1]The Graduate University for Advanced Studies, SOKENDAI, Tachikawa, Tokyo 190-8562, Japan
[2]The Institute of Statistical Mathematics, Research Organization of Information and Systems, Tachikawa, Tokyo 190-8562, Japan
[3]National Institute for Materials Science, Tsukuba, Ibaraki 305-0047, Japan
[4]National Institute of Advanced Industrial Science and Technology, Tsukuba, Ibaraki 305-8560, Japan

[*]kusaba@ism.ac.jp, yoshidar@ism.ac.jp

## ABSTRACT

In 1869, the first draft of the periodic table was published by Russian chemist Dmitri Mendeleev. In terms of data science, his achievement can be viewed as a successful example of feature embedding based on human cognition: chemical properties of all known elements at that time were compressed onto the two-dimensional grid system for a tabular display. In this study, we seek to answer the question of whether machine learning can reproduce or recreate the periodic table by using observed physicochemical properties of the elements. To achieve this goal, we developed a periodic table generator (PTG). The PTG is an unsupervised machine learning algorithm based on the generative topographic mapping (GTM), which can automate the translation of high-dimensional data into a tabular form with varying layouts on-demand. The PTG autonomously produced various arrangements of chemical symbols, which organized a two-dimensional array such as Mendeleev's periodic table or three-dimensional spiral table according to the underlying periodicity in the given data. We further showed what the PTG learned from the element data and how the element features, such as melting point and electronegativity, are compressed to the lower-dimensional latent spaces.

## Introduction

The periodic table is a tabular arrangement of elements such that the periodic patterns of their physical and chemical properties are clearly understood. The prototype of the current periodic table was first presented by Mendeleev in 1869 [1]. At that time, about 60 elements and their few chemical properties were known. When the elements were arranged according to their atomic weight, Mendeleev noticed an apparent periodicity and an increasing regularity. Inspired by this discovery, he constructed the first periodic table. Despite the subsequent emergence of significant discoveries [2, 3], including the modern quantum mechanical theory of the atomic structure, Mendeleev's achievement is still the de facto standard. Regardless, the design of the periodic table continues to evolve, and hundreds of periodic tables have been proposed in the last 150 years [4, 5]. The structures of these proposed tables have not been limited to the two-dimensional tabular form, but also spiral, loop, or three-dimensional pyramid forms [6, 7, 8].

The periodic tables proposed so far have been products of human intelligence. However, a recent study has attempted to redesign the periodic table using computer intelligence—machine learning [9]. From this approach, building a periodic table can be viewed as an unsupervised learning task. Precisely, the observed physicochemical properties of elements are mapped onto regular grid points in a two-dimensional latent space such that the configured chemical symbols adequately capture the underlying periodicity and similarity of the elements. Lemes & Pino [9] used Kohonen's self-organizing map (SOM) [10] to place five-dimensional features of elements (i.e. atomic weight, radius of connection, atomic radius, melting point, and reaction with oxygen) into two-dimensional rectangular grids. This method successfully placed similarly behaved elements into neighbouring sub-regions in the lower-dimensional spaces. However, the machine learning algorithms never reached Mendeleev's achievement as they missed important features such as between-group and between-family similarities.

In this study, we created various periodic tables using a machine learning algorithm. The dataset that we used consisted of 39 features (melting points, electronegativity, and so on) of 54 elements with the atomic number 1-54, corresponding to hydrogen to xenon (Fig. S1 for the heatmap display). A wide variety of dimensionality reduction methods has so far been made available, such as principal component analysis (PCA), kernel PCA [11], isometric feature mapping (ISOMAP) [12], local linear embedding (LLE) [13], and t-distributed stochastic neighbour embedding (t-SNE) [14]. However, none of these methods could well visualize underlying periodic laws (Supplementary Fig. S3). To begin with, none of these methods offers a tabular representation. The task of building a periodic table can be regarded as the dimension reduction of the element data to arbitrary given ‘discrete’ points rather than a continuous space. To the best of our knowledge, no existing framework is available for such table summarization tasks. Therefore, we developed a new unsupervised machine learning algorithm called the periodic table generator (PTG), which relies on the generative topographic mapping (GTM) [15] with latent variable dependent length-scale and variance (GTM-LDLV) [16]. With this, we can automate the process of translating patterns of high-dimensional feature vectors to an arbitrary given layout of lower dimensional point clouds.

The PTG produced various arrangements of chemical symbols, which organized, for example, a two-dimensional array such as Mendeleev’s table or three-dimensional spiral table according to the underlying periodicity in the given data. We will show what the machine intelligence learned from the given data and how the element features were compressed to the reduced dimensionality representations. The periodic tables can also be regarded as the most primitive descriptor of chemical elements. Hence, we will highlight the representation capability of such element-level descriptors in the description of materials that were used in machine learning tasks of materials property prediction.

## Materials and Methods

### Computational workflow

The workflow of the PTG begins by specifying a set of point clouds, called ‘nodes’ hereafter, in a low-dimensional latent space to which chemical elements with observed physicochemical features are assigned. The nodes can take any positional structure such as equally spaced grid points on a rectangular for an ordinal table, spiral, cuboid, cylinder, cone, and so on. A Gaussian process (GP) model [17] is used to map the pre-defined nodes to the higher-dimensional feature space in which the element data are distributed. A trained GP defines a manifold in the feature space to be fitted with respect to the observed element data. The smoothness of the manifold is governed by a specified covariance function called the kernel function, which associates the similarity of nodes in the latent space with that in the feature space. The estimated GP defines a posterior probability or responsibility of each chemical element belonging to one of the nodes. An element is assigned to one node with the highest posterior probability.

As indicated by the failure of some existing methods of statistical dimension reduction, such as PCA, t-SNE, and LLE, the manifold surface of the mapping from chemical elements to their physiochemical properties is highly complex. Therefore, we adopted the GTM-LDLV as a model of PTG, which is a GTM that can model locally varying smoothness in the manifold. To ensure non-overlapping assignments such that no multiple elements shared the same node, we operated the GTM-LDLV with the constraint of one-to-one matching between nodes and elements. To satisfy this, the number of nodes, $K$, has to be larger than the number of elements, $N$. However, a direct learning with $K > N$ suffers from high computational costs and instability of the estimation performance. Specifically, the use of redundant nodes leads to many suboptimal solutions corresponding to undesirable matchings to the chemical elements. To alleviate this problem, the PTG was designed to take a three-step procedure (Fig. 1) that relies on a coarse-to-fine strategy. In the first step, we operated the training of GTM-LDLV with a small set of nodes such that $K < N$. In the following step, we generated additional nodes such that $K > N$, and the expanded node-set was transferred to the feature space by performing the interpolative prediction made by the given GTM-LDLV. Finally, the pre-trained model was fine-tuned subject to the one-to-one matching between the $N$ elements and the $K$ nodes for tabular construction. The procedure for each step is detailed below.

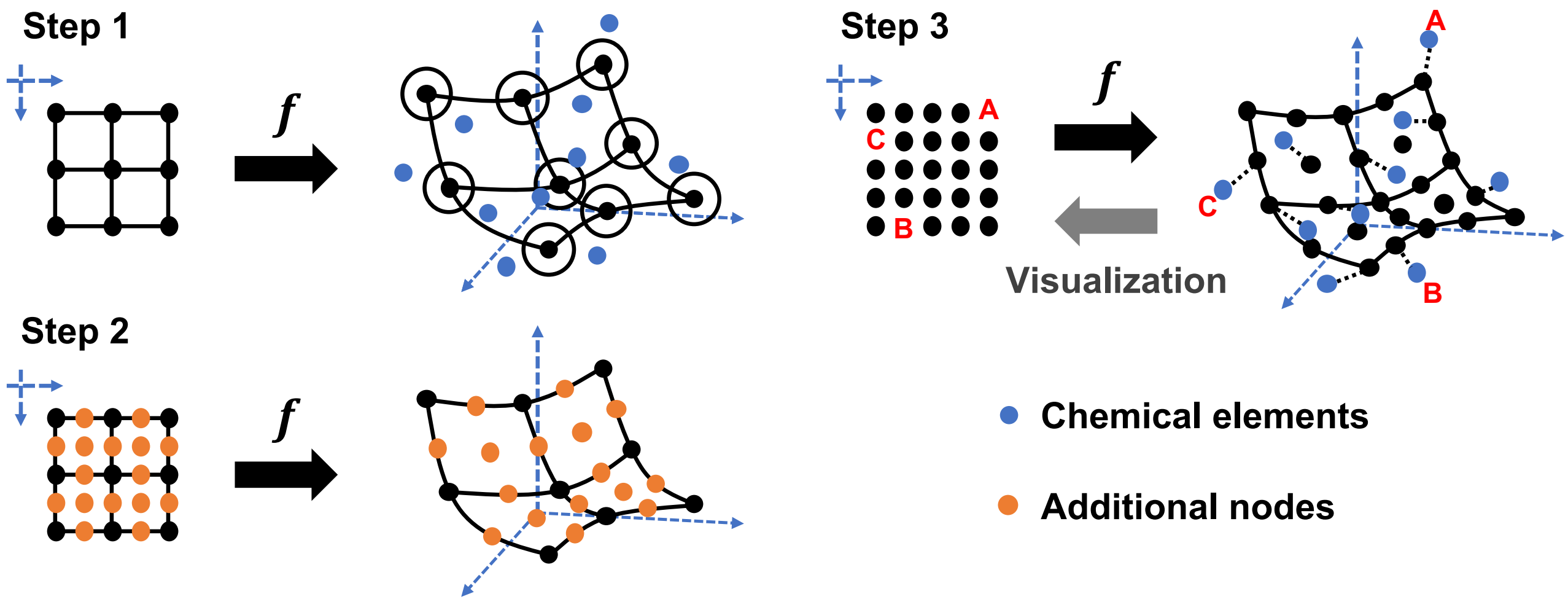

**Figure 1.** Workflow of PTG that relies on a three-step coarse-to-fine strategy to reduce the occurrence of undesirable matching between chemical elements and redundant nodes.

*Step 1* (GTM-LDLV): The first step of the PTG is the same as the original GTM-LDLV. In the GTM-LDLV, $K$ nodes, $\boldsymbol{u}_1, \cdots, \boldsymbol{u}_K$, arbitrarily arranged in the $L$-dimensional latent space are first prepared. Then we build a nonlinear function $\boldsymbol{f}(\boldsymbol{u}_k)$ that maps the pre-defined nodes to the $D$-dimensional feature space. The model $\boldsymbol{f}(\boldsymbol{u}_k)$ defines an $L$-dimensional manifold in the $D$-dimensional feature space, which is fitted with respect the $N$ data points of element features. The dimension of the latent space is set to $L \leq 3$ for visualization.

It is assumed that the $D$-dimensional feature vector $\boldsymbol{x}_n$ of element $n$ is generated independently from a mixture of $K$ Gaussian distributions, where the mixing rates are all equal to $1/K$, and the mean and the covariance matrix of each distribution are $\boldsymbol{y}_k = \boldsymbol{f}(\boldsymbol{u}_k)$ and $\beta^{-1}\boldsymbol{I}$, respectively ($\boldsymbol{I}$ denotes the identity matrix). According to the GTM-LDLV, the mean $\boldsymbol{f}(\boldsymbol{u}_k)$ is modelled to be the product of two functions, a $D$-dimensional vector-valued function $\boldsymbol{h}(\boldsymbol{u}_k)$ and a positive scalar function $g(\boldsymbol{u}_k)$. Here, we introduce a vector of $K$ latent variables, $\boldsymbol{z}_n = (z_{1n}, \cdots, z_{Kn})'$, that indicates the assignment of element $n$ to one of the given $K$ nodes. The $k$th entry $z_{kn}$ takes the value of 1 if $\boldsymbol{x}_n$ is generated by the $k$th component distribution, and 0 otherwise. Here, let $\boldsymbol{X}$ denote a matrix of $\boldsymbol{x}_1, \cdots, \boldsymbol{x}_N$ of the elements, and $\boldsymbol{Z}$ be a matrix of $\boldsymbol{z}_1, \cdots, \boldsymbol{z}_N$. Then, their joint distribution is given by

$$p(\boldsymbol{X}, \boldsymbol{Z} | \boldsymbol{g}, \boldsymbol{H}, \beta) = K^{-N} \prod_{n=1}^{N} \prod_{k=1}^{K} N(\boldsymbol{x}_n | \boldsymbol{y}_k, \beta^{-1}\boldsymbol{I})^{z_{kn}}, \tag{1}$$

$$\boldsymbol{y}_k = \boldsymbol{f}(\boldsymbol{u}_k) = g(\boldsymbol{u}_k)\boldsymbol{h}(\boldsymbol{u}_k), \tag{2}$$

where $N(\cdot | \boldsymbol{\mu}, \boldsymbol{\Sigma})$ denotes the Gaussian density function with mean $\boldsymbol{\mu}$ and covariance matrix $\boldsymbol{\Sigma}$, $\boldsymbol{g}$ is a vector of $g(\boldsymbol{u}_k)$ $(k = 1, \cdots, K)$, and $\boldsymbol{H}$ is a matrix of $\boldsymbol{h}(\boldsymbol{u}_k)$ $(k = 1, \cdots, K)$.

The prior distribution of $g(\boldsymbol{u})$ is given as a truncated GP with mean 0 and covariance function $c_g(\boldsymbol{u}_i, \boldsymbol{u}_j; \boldsymbol{\xi}_g)$, which handles positive-bounded random functions. The prior distribution of the $d$th entry $h_d(\boldsymbol{u})$ of $\boldsymbol{h}(\boldsymbol{u})$ is given as a GP with mean 0 and covariance function $c_h(\boldsymbol{u}_i, \boldsymbol{u}_j)$. To be specific, the covariance functions, $c_g(\boldsymbol{u}_i, \boldsymbol{u}_j; \boldsymbol{\xi}_g)$ and $c_h(\boldsymbol{u}_i, \boldsymbol{u}_j)$, are given by

$$c_g(\boldsymbol{u}_i, \boldsymbol{u}_j; \boldsymbol{\xi}_g) = \nu_g \cdot \exp\left(-\frac{\|\boldsymbol{u}_i - \boldsymbol{u}_j\|^2}{2l_g}\right), \tag{3}$$

$$c_h(\boldsymbol{u}_i, \boldsymbol{u}_j) = \left\{\frac{2l(\boldsymbol{u}_i)l(\boldsymbol{u}_j)}{l^2(\boldsymbol{u}_i) + l^2(\boldsymbol{u}_j)}\right\}^{\frac{L}{2}} \exp\left(-\frac{\|\boldsymbol{u}_i - \boldsymbol{u}_j\|^2}{l^2(\boldsymbol{u}_i) + l^2(\boldsymbol{u}_j)}\right). \tag{4}$$

In equation (3), the hyperparameter $\boldsymbol{\xi}_g$ consists of $\nu_g$ and $l_g$, referred to as the variance and the length-scale, that control the magnitude of variances and smoothness of a positive-valued function $g(\boldsymbol{u})$ generated from the GP. In equation (4), the length-scale parameter $l(\boldsymbol{u})$ is a function of $\boldsymbol{u}$ and parameterized as $l(\boldsymbol{u}) = \exp(r(\boldsymbol{u}))$ with the function $r(\boldsymbol{u})$ following the GP with mean 0 and covariance function $c_r(\boldsymbol{u}_i, \boldsymbol{u}_j; \boldsymbol{\xi}_r)$. Finally, a gamma prior is placed on the precision parameter $\beta$ in equation

(1).

The covariance function in equation (4) is the key in the GTM-LDLV. In general, a covariance function in a GP governs a degree of preservation between the similarity of any inputs, e.g. $\boldsymbol{u}_i$ and $\boldsymbol{u}_j$, and the similarity of their outputs. The heterogeneous variance over the latent space in equation (4) can bring locally varying smoothness in resulting manifolds in the feature space. In addition, the variance function is statistically estimated with the hierarchically specified GP prior based on the covariance function $c_r(\boldsymbol{u}_i, \boldsymbol{u}_j; \boldsymbol{\xi}_r)$.

The unknown parameter to be estimated is $\boldsymbol{\theta} = \{\boldsymbol{Z}, \beta, \boldsymbol{g}, \boldsymbol{H}, \boldsymbol{r}\}$. In the GTM-LDLV, the posterior distribution $p(\boldsymbol{\theta}|\boldsymbol{X})$ is approximately evaluated using a Markov Chain Monte Carlo (MCMC) method. Iteratively sampling from the full conditional posterior distribution for each $\{\boldsymbol{Z}, \beta, \boldsymbol{g}, \boldsymbol{H}, \boldsymbol{r}\}$, we obtained a set of ensembles that follow the posterior distribution approximately. By taking the ensemble average over the samples from $p(\boldsymbol{\theta}|\boldsymbol{X})$, the parameters of the GTM-LDLV are estimated. A detailed description of the GTM-LDLV is given in the Supplementary Information section.

*Step 2* (node expansion): To avoid the occurrence of improper assignments of the $N$ elements to a redundant set of nodes, we adopt a coarse-to-fine strategy. Starting from an initially trained GP model of $K < N$ at step 1, we refine the model with an increased number of nodes $K \geq N$. For example, $5 \times 5$ nodes evenly arranged on the area $[-1, 1] \times [-1, 1]$ at step 1 are incremented to $K = 9 \times 9$ by placing additional nodes at middle points of the line segments connecting between each node. With the currently given parameters, we can infer the values of $r(\boldsymbol{u})$ of the covariance function in equation (4) at the expanded nodes, $\boldsymbol{u}_1, \cdots, \boldsymbol{u}_K$. Likewise, the values of $g(\boldsymbol{u})$ and $\boldsymbol{h}(\boldsymbol{u})$ are interpolated. By performing such initialization, we proceed to the next round of the GTM-LDLV.

*Step 3* (GTM-LDLV subject to one-to-one assignments): Finally, the resulting GTM-LDLV is fine-tuned to obtain a tabular display by running the above procedure subject to a one-to-one matching between the $N$ elements and the $K$ nodes. By definition, the conditional posterior distribution of the assignment variables is represented as

$$p(\boldsymbol{Z}|\boldsymbol{X}, \boldsymbol{\theta}_{-\boldsymbol{Z}}) \propto \prod_{n=1}^{N} \prod_{k=1}^{K} \exp\left(-\frac{\beta}{2} \|\boldsymbol{x}_n - \boldsymbol{y}_k\|^2\right)^{z_{kn}} = \exp\left(-\frac{\beta}{2} \sum_{n=1}^{N} \sum_{k=1}^{K} z_{kn} \|\boldsymbol{x}_n - \boldsymbol{y}_k\|^2\right),$$

where $\boldsymbol{\theta}_{-\boldsymbol{A}}$ represents a set of the parameters obtained by removing $\boldsymbol{A}$ from $\boldsymbol{\theta}$**.** In the MCMC calculation in step 1, we iteratively draw a sample of $\boldsymbol{Z}$ from this distribution. Here, instead of performing the random sampling, we conduct the maximization of the logarithmic posterior with respect to $\boldsymbol{Z}$ subject to the constraint of one-to-one assignments. The problem amounts to finding the solution of

$$\max_{\boldsymbol{Z} \in A} - \sum_{n=1}^{N} \sum_{k=1}^{K} z_{kn} \|\boldsymbol{x}_n - \boldsymbol{y}_k\|^2,$$

$$A = \left\{\boldsymbol{Z} \middle| \sum_{k=1}^{K} z_{kn} = 1 \ (n = 1, \cdots, N), \quad \sum_{n=1}^{N} z_{kn} \leq 1 \ (k = 1, \cdots, K)\right\}.$$

This is regarded as a transportation problem where the sum of the squared Euclidean distance between an element feature $\boldsymbol{x}_n$ and a node $\boldsymbol{y}_k$ embedded in the feature space is the cost of transporting one item from source $k$ to destination $n$ under the constraint $A$. We use the lpSolve package [18] in R [19] to solve the transportation problem.

This partially modified MCMC is iterated few times (e.g. $T = 10$) to make a fine-tuning of the currently given parameters. The assignment variables and the other parameters that exhibit the highest likelihood are chosen to form the final estimate of sthe PTG. A summary of the algorithm of PTG is shown in Supplementary Algorithm 1.

### Interpretation

The PTG autonomously creates a tabular display of the chemical elements according to the estimated $\boldsymbol{Z}$. To understand how the element features such as melting point and electronegativity are compressed on the low-dimensional tabular display, each of the features is mapped onto the resulting table. Specifically, we overlay a smoothed heatmap of each feature on the table. With this PTG property landscape [20], we can visually understand the distribution of the topographical mapping that indicates how the element features are embedded in the latent space.

### Periodic table as an element descriptor

We consider an evaluation basis for the quality of a designed periodic table in terms of a novel view from data science. A periodic table, including Mendeleev's classic table, can be considered as one of the most primitive descriptors that encodes known element features into the coordinate system of a low-dimensional latent space. Neighbouring elements on a table should behave similarly and possess similar physicochemical properties. Inspired by such an idea, we consider the use of a periodic table as a descriptor of chemical elements in a task of predicting materials properties based on machine learning [21]. The periodic table is then evaluated quantitatively based on the predictive performance of the descriptor.

For a given table, its coordinates $\boldsymbol{u}_{k(1)}, \cdots, \boldsymbol{u}_{k(N)}$ of the nodes to which the $N$ elements are assigned are used as a set of element descriptors. For a compound $S$, its fraction of the $N$ elements is denoted by $w_1(S), \cdots, w_N(S)$ where $0 \leq w_n(S) \leq 1$ and $\sum_{n=1}^{N} w_n(S) = 1$. The compositional descriptor of $S$ is calculated by $\boldsymbol{\phi}(S) = \sum_{n=1}^{N} w_n(S)\, \boldsymbol{u}_{k(n)}$. With this descriptor, we derive a prediction model $Y = f(\boldsymbol{\phi}(S))$, which is trained in $m$ training instances $\{Y_i, S_i\}_{i=1}^{m}$, that describes a physicochemical property $Y$ as a function of the descriptor $\boldsymbol{\phi}(S)$ for any given compound $S$. Descriptors exhibiting higher predictability should be recognised as providing more efficient compression performances on the $N$ elements.

### Data: element features

The element feature set was extracted from XenonPy [22], which is a Python library for materials informatics, by using an Application Programming Interface (API) (see the XenonPy website [23]). The original dataset consisted of 74 features of 118 elements. Since elements with large atomic numbers contained many missing values, we selected 54 elements with the atomic number 1-54 corresponding to hydrogen to xenon that are considered sufficient to retain the periodic rule. After removing features that contained one or more missing values, the dataset was reduced to 39 features of 54 elements. For the $54 \times 39$ data matrix, each feature (column) was standardized to have mean 0 and variance 1. A heatmap display of the data matrix and a detailed description of the 39 features are provided in Supplementary Figs. S1 and S2, respectively.

### Analysis procedure

We performed the PTG on two different layouts of nodes, square, and three-dimensional conical layouts. In the square layout of $L = 2$, we set $K = 25$ in the first step of PTG in which the $5 \times 5$ nodes were evenly arranged on the area $[-1, 1] \times [-1, 1]$. In the second step, we increased the number of nodes to $9 \times 9$ by placing new nodes at the middle points of the line segments connecting between each node. In the conical layout of $L = 3$, we first used a set of nodes with $K = 25$ that were arranged uniformly on the surface of the cone placed in the area $[-1, 1] \times [-1, 1] \times [-1, 1]$. The cone was sliced into 4 sections in the same height along the vertical axis. Then, 1 (vertex), 4, 8, and 12 (bottom) nodes were uniformly placed on the outer part of the 4 cut surfaces. In the next step, the number of slices was increased by 7, and 1 (vertex), 4, 8, 12, 16, 20, and 24 (bottom) nodes were uniformly arranged in the same way. In both the cases, we set $\boldsymbol{\xi}_g = \boldsymbol{\xi}_r = (1/3, 3)$, the number of iteration in MCMC was set to $T = 10{,}000$ with the burn-in step $T_b = 5{,}000$, and the number of iteration in the third step of fine-tuning was set to $T = 10$. See the Supplementary Information section for further details on the hyperparameter settings and analysis procedure.

The PTG algorithm was implemented using R codes, which are available at [24] with the element dataset. Readers can run the PTG algorithm with the element data used in this paper.

## Results

### Results of PTG

***Square table***

Figure 2 shows a PTG-created layout of the 54 elements on the $9 \times 9$ square lattice. Elements in each period of the standard periodic table were configured in a fan shape from the top left to the bottom right. The order of atomic numbers was also almost reconstructed in the PTG table. The elements in the square table are clearly separated into metal and nonmetal by the red dashed line shown in Fig. 2. The 3d and 4d transition elements were separated and both clustered in the lower right. In addition, the elements were clearly clustered by groups such as alkali metals, alkaline earth metals, halogens, and noble gases. This looked like a variant of the original periodic table: the original table was folded around the centre on which transition elements are positioned, the two separated blocks of group 1-2 and 13-18 in the first to third periods were brought nearer with each other while keeping away from the area of transition elements, and they were stored into the square table. Notably, the square table exhibited the discontinuity from group 18 to group 1 as in the original table. Though results are not shown, the same discontinuity appeared frequently in most square tables created in the experiments under different conditions.

***Conical table***

Figure 3 shows a layout on the three-dimensional conical nodes. The elements were arranged in a spiral structure starting from the top of the cone according to increasing atomic numbers. Viewed from the top, the elements were stratified concentrically by the periods of the standard periodic table. This view was slightly similar to the circular periodic table that was constructed in a different study [7]. One block corresponded to a set of elements divided according to the orbital type of the electrons of the highest energy levels. In the standard periodic table, helium (He: an element circled by the red line in Fig. 4) is located away from the other s-block elements (a set of elements coloured red in Fig. 4), but in the conical table, it was located close to them. It was also seen that the elements in the conical table were clearly classified into typical elements and transition elements by the red line shown in Fig. 4. A blank space was observed between group 1 and group 18 on the conical table implying that there is a gap of properties between them in the feature space.

In the spiral structure viewed from above, the atomic numbers were monotonically arranged from top to bottom except for a few elements. The disorder appeared in group 6 to 7: chromium (Cr: atomic number = 24) and manganese (Mn: 25) in period 4 or molybdenum (Mo: 42) and technetium (Tc: 43) in period 5. In both the square and conical tables, the elements were arranged radially according to groups, and elements of group 1 and 2 were located a little away from group 3.

## Interpretation

To understand how the element features have been embedded on the created tables, each of the features was mapped on the lower-dimensional latent space (Fig. 5). In the property landscape of the conical table, atomic radius increased gradually and concentrically from the top of the cone, electron negativity decreased gradually and concentrically from the top of the cone, and melting point gradually increased from right to left. The distribution of thermal conductivity looked a little more complicated than the former three, but continuity and unimodality still held on the surface of the three-dimensional conical table. On the other hand, in the square table, the landscapes of some element features, e.g. atomic radius and thermal conductivity, exhibited multimodality. This discontinuity arose from the unnatural layout of the elements in the two-dimensional tabular representation as in the standard periodic table. The PTG property landscapes of the 39 features are shown in the Supplementary Information section.

## Quantitative comparison of periodic tables

To evaluate the validity of a periodic table and uncover the information gain and loss of the reduced representation, we considered the use of a table as an element descriptor in machine learning tasks. The task to be addressed was the prediction of formation energies of inorganic compounds. The dataset that we used for the training of random forest regressors (RF) [25] was obtained from Materials Project [26], which is a database of materials properties generated from high-throughput first-principles calculations. Among all inorganic compounds in Materials Project, we selected compounds that are stable and consist of elements with the atomic number 1-54 (H to Xe). The dataset consisted of the formation energies per atom of 12,373 inorganic compounds.

The objective here was to train an RF that describes the formation energy as a function of the conical descriptor $\boldsymbol{\phi}(S)$ obtained by composing $S$ and the three-dimensional coordinates of the elements in the conical table. This is described in the Methods section above. For comparison, we built models using two descriptors based on the two-dimensional coordinates in the created square table and the standard periodic table, similar to the conical descriptor.

We performed the five-fold cross-validation on the 12,373 samples for the three types of descriptors. As shown in Fig. 6, the conical PTG achieved a mean absolute error (MAE) of 0.464 eV/atom and a root mean square error (RMSE) of 0.643 eV/atom, whereas the MAE and the RMSE of the square PTG and the standard periodic table were 0.533 eV/atom and 0.719 eV/atom, and 0.549 eV/atom and 0.734 eV/atom, respectively. In summary, the square PTG was slightly superior to the standard periodic table, but the conical PTG table outperformed the standard periodic table and the square PTG.

A detailed investigation of the prediction results provided some insights into the difference in information compression between the three-dimensional conical table and the standard periodic table. We focused on a subset of the compounds used in the validation, hereafter denoted by $D_{\mathrm{cone}}$ (i.e. the conical descriptor dominant set), that had the MAE values less than 0.3 eV/atom for the conical descriptor, but 1.0 eV/atom greater than the conical descriptor for the standard periodic table. Likewise, we identified $D_{\mathrm{standard}}$ with the MAE values less than 0.3 eV/atom for the standard periodic table, but 1.0 eV/atom greater than the standard periodic table for the conical table. We counted the frequency of a chemical element in $D_{\mathrm{cone}}$ and $D_{\mathrm{standard}}$, and evaluated the enrichment of the element by comparing its expected frequency calculated with the background, i.e. the number of occurrence in the overall population (the 12,373 compounds in Materials Project). As shown in Fig. 7, a significantly enriched group in $D_{\mathrm{cone}}$ comprised transition elements in the fourth period that correspond to atomic number 21-29. Aluminium (Al) was also enriched in $D_{\mathrm{cone}}$ (Fig. 7: a set of elements circled by a blue line). Notably, these over-represented elements formed a cluster in the created conical table (Fig. 4: a set of elements circled by a blue line). On the other hand, hydrogen (H) was significantly enriched in $D_{\mathrm{standard}}$ (Fig. 7: an element circled by green line). H is located just above lithium (Li) in the standard periodic table (Fig. 4: an element circled by a green line), while it was located between fluorine (F) and Li in the conical periodic table.

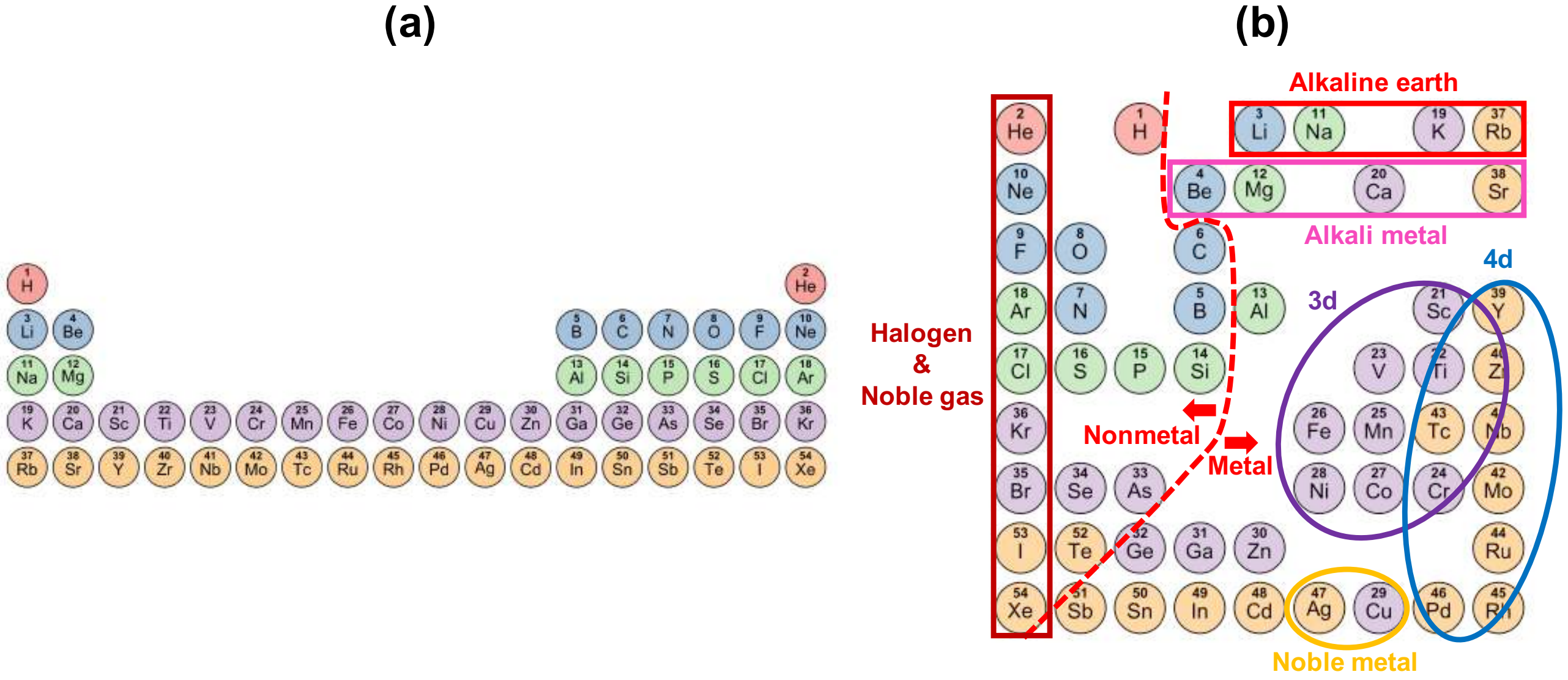

**Figure 2.** (a) The currently most common periodic table of the elements. (b) Square PTG table created from the training data of 39 features of the 54 elements. The elements are colour-coded by periods and numbered by atomic numbers.

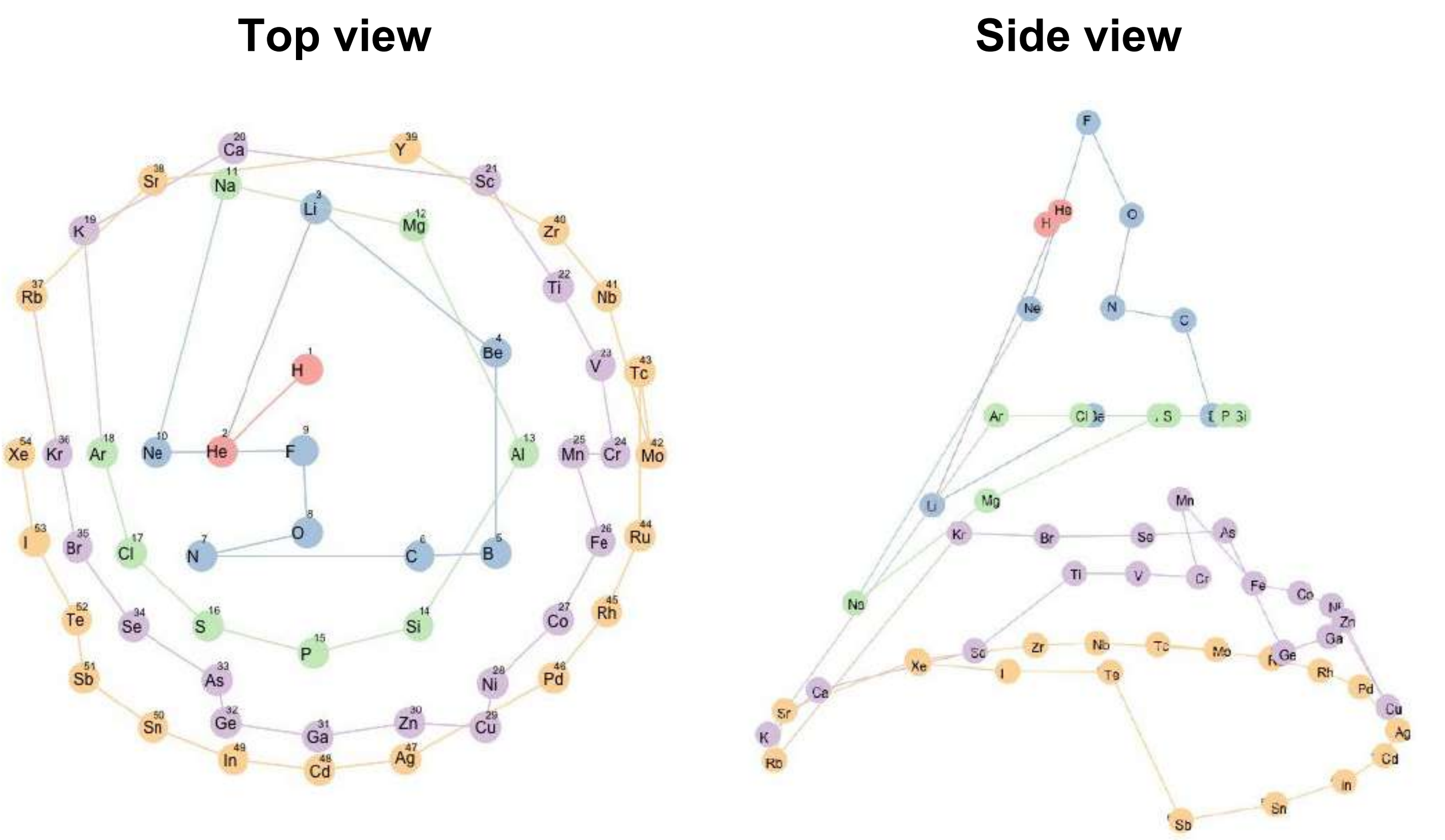

**Figure 3.** Created conical table of 54 chemical elements. The elements are colour-coded according to five periods and numbered by atomic number. A line passing through the elements is drawn in the order of atomic numbers. The left and right figures show the same table viewed from top and side, respectively.

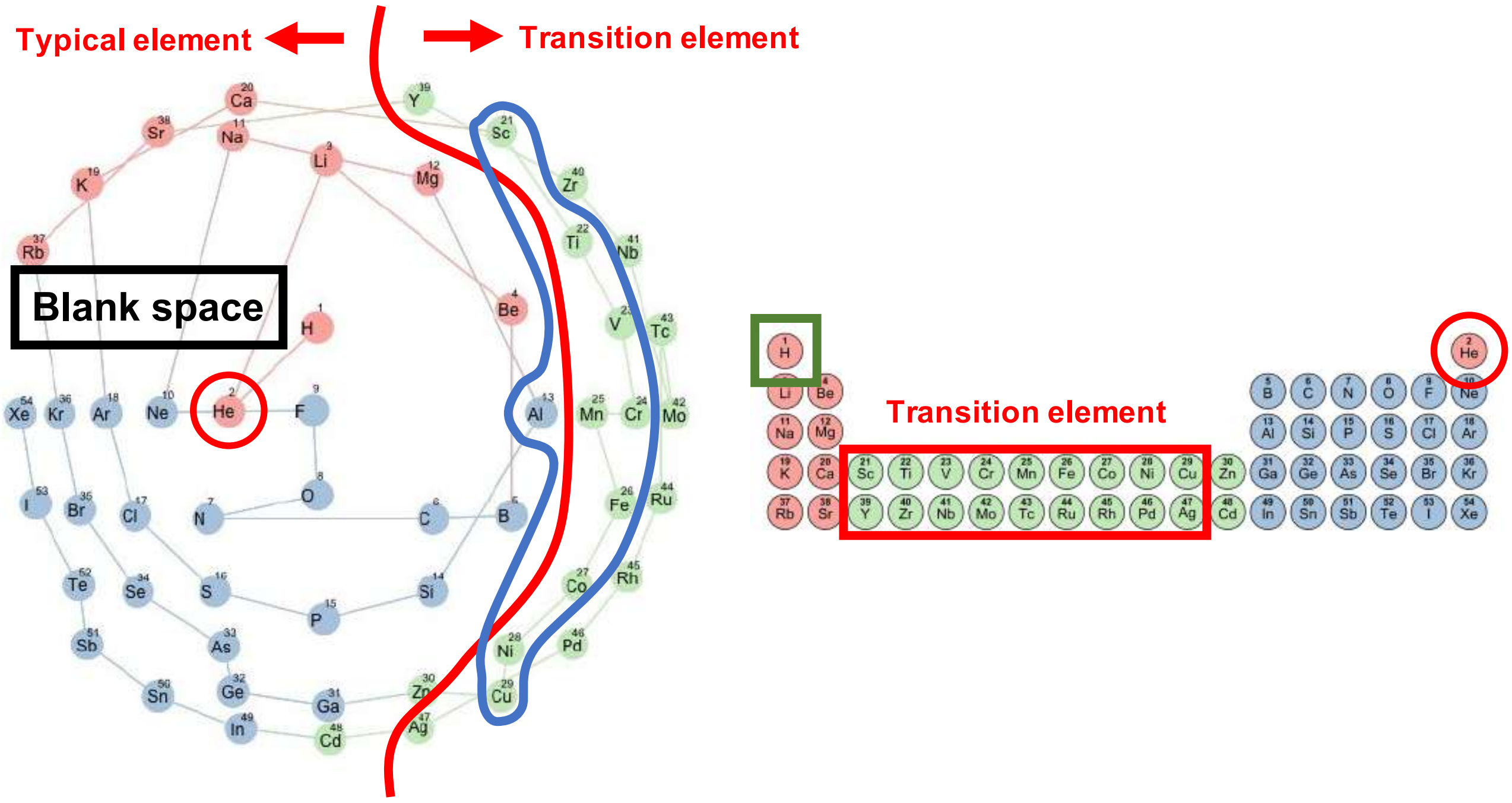

**Figure 4.** The left panel shows a conical table viewed from above. The elements are colour-coded according to three blocks in the standard periodic table that are indicated in the right panel. The red line in the left indicates the segment between transition

elements and typical elements.

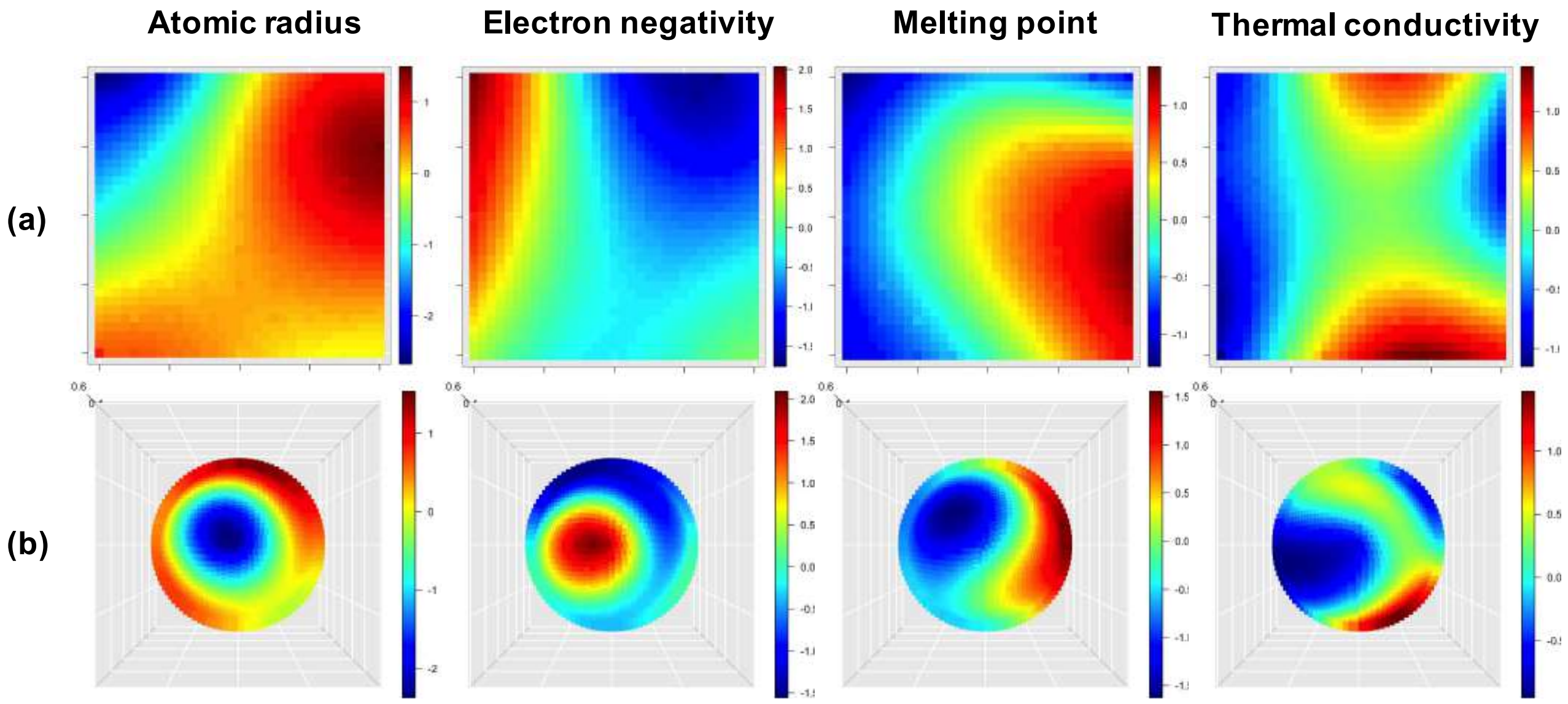

**Figure 5.** Property landscapes of atomic radius (Rahm et al. [27]), electron negativity, melting point, and thermal conductivity at 25°C that are embedded in the latent spaces. The heatmaps are laid on (a) the square table in Fig. 2 and (b) the conical table (top view) in Fig. 3.

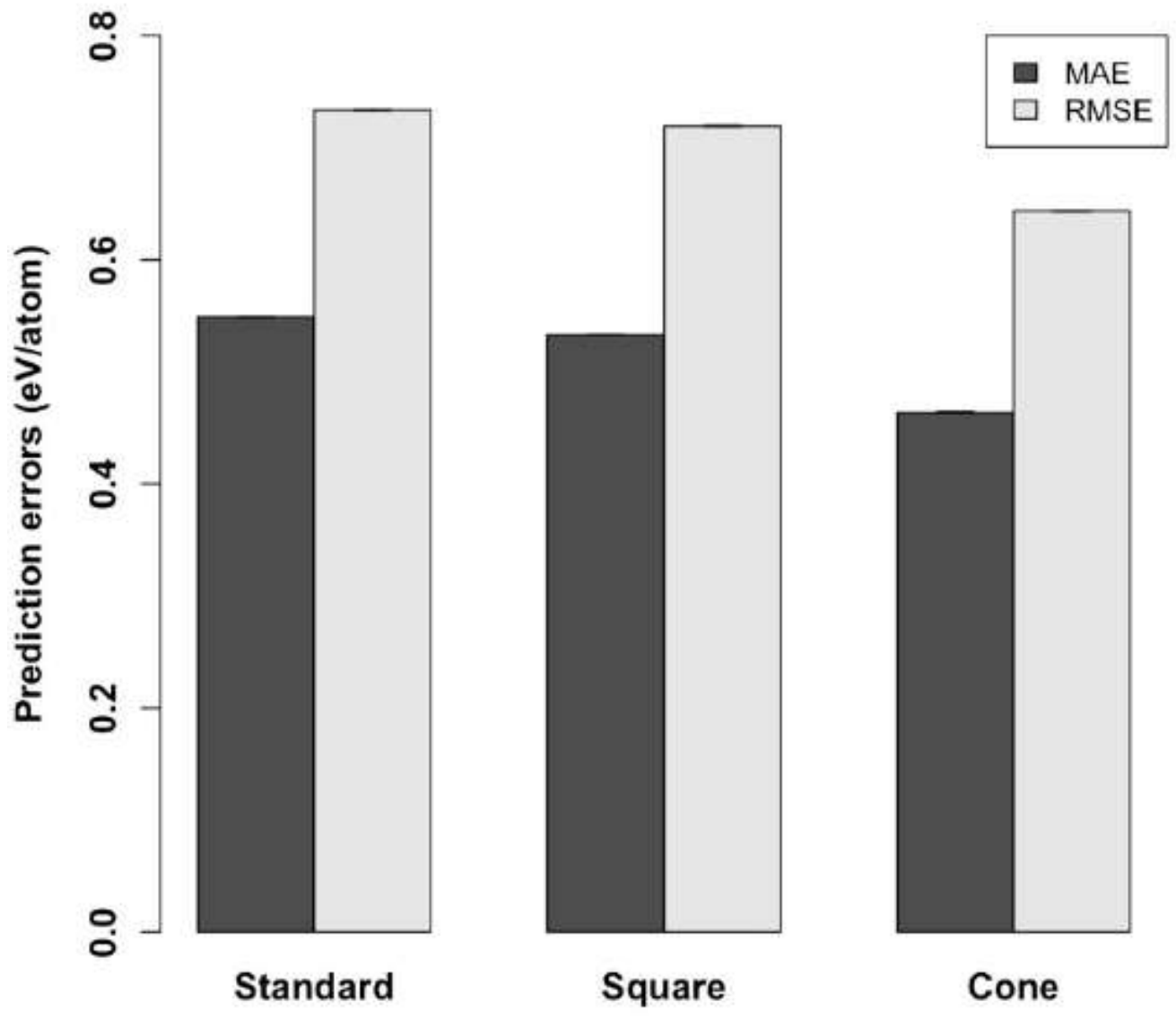

**Figure 6.** Performance of the prediction of the formation energy per atom for three periodic tables used as element descriptors. The vertical axis indicates cross-validated MAE and RMSE of RF regressors trained with the three descriptors obtained from the coordinates of elements in the standard periodic table (left), the square PTG table (middle), and the conical PTG table (right), respectively. The error bars denote the standard deviations in five independent trials of the cross-validation (the error bars are invisible because of substantially small scales).

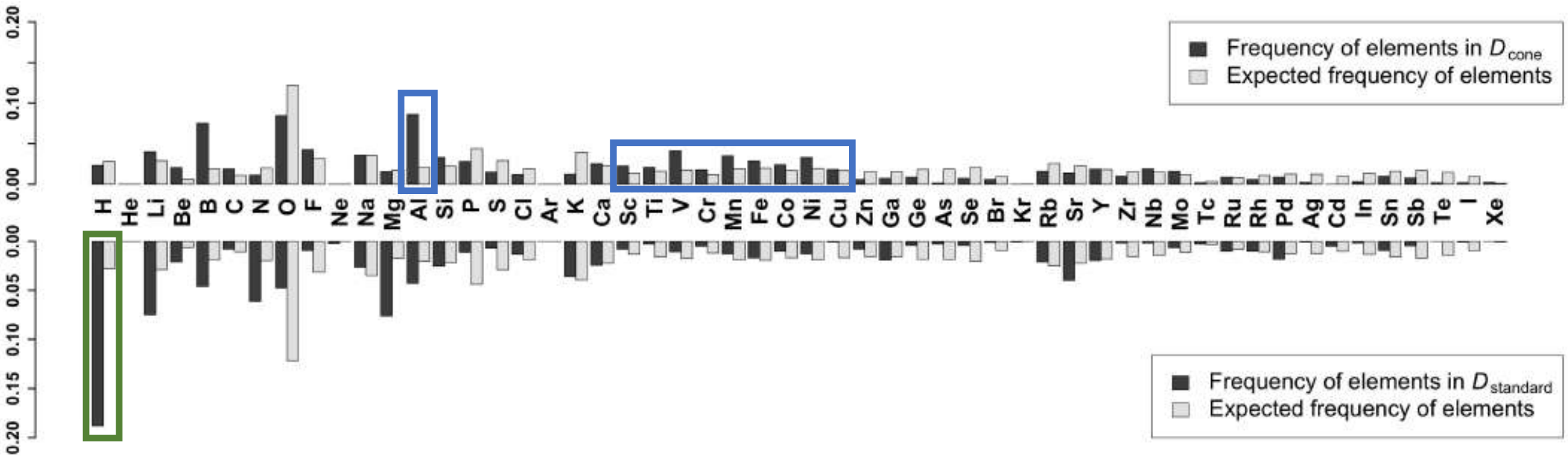

**Figure 7.** Comparison of the frequencies of chemical elements in $D_{\mathrm{cone}}$ (top: black bar chart) and $D_{\mathrm{standard}}$ (bottom: black bar chart). White bar charts show the expected frequency calculated with the number of occurrences in the overall population.

## Concluding Remarks

Since the emergence of Mendeleev's periodic table, hundreds of redesigned tables have been created. In terms of machine learning, the tabular construction can be considered a task of reducing the dimensionality of high-dimensional data. A previous study first attempted to yield the periodic table using machine learning by applying SOM to five element features available in Mendeleev's time [9]. Though the SOM successfully placed similarly behaved elements in neighbouring sub-regions on the table, the reported results still never reached Mendeleev's achievement as it obviously failed to capture the underlying periodicity of the elements. To reach Mendeleev's achievement, we attempted to develop PTG as an unsupervised machine learning algorithm that can automate the translation of high-dimensional data into a tabular form with varying layouts on-demand.

In the previous study based on SOM, some chemical elements having similar properties occupied the same cell in the table due to SOM inability to guarantee non-overlapping assignments of elements. When we began this study, there had been no existing machine learning methods for the task of tabular construction. To the best of our knowledge, the PTG algorithm that we present is the first tabular constructor based on machine learning, yet this is a secondary contribution of this study.

In this study, we created the two types of periodic tables. The square table was considerably similar to the currently most common periodic table, but some outstanding differences were observed, for example in the arrangement of H and He. These elements were placed far away in the standard periodic table, but their physicochemical properties were similar. The PTG suggested that these elements should be put closer according to the observed data. The three-dimensional layout on the cone also provided some insight into how the transition elements in the fourth period, including aluminium (Al), should be arranged. In addition, the created conical table provided a re-ordering from Cr to Mn in period 4 and from Mo to Tc in period 5 in the standard table.

A periodic table is the most basic descriptor of chemical elements. Historically, the primary design objective has focused on the understandability and the interpretability to humans even at the expense of reducing some key detailed features. Here, we provided a new way of looking at periodic tables. The coordinates of elements put on a table can be considered as an element descriptor, which is also converted to a descriptor of materials. The quality of designed tables should be assessed on the performance of predicting physicochemical properties of resulting machine learning models. This study focused only on the prediction of formation energies, but more diverse properties should be incorporated into the design objective. Also, we focused only on the two types of layouts, but there are a lot of options for potentially promising layouts. Our algorithm would contribute to the recreation of more sophisticated tabular displays of chemical elements.

## Acknowledgements

This work was supported in part by the Materials Research by Information Integration Initiative (MI2I) of the Support Program for Starting Up Innovation Hub from the Japan Science and Technology Agency (JST). Ryo Yoshida acknowledges financial support from a Grant-in-Aid for Scientific Research (A) 19H01132 from the Japan Society for the Promotion of Science (JSPS), JST CREST Grant Number JPMJCR19I3, Japan, and JSPS KAKENHI Grant Number 19H50820.

## Author Contributions

Minoru Kusaba and Ryo Yoshida designed the research; Minoru Kusaba and Ryo Yoshida wrote the manuscript; Minoru Kusaba wrote the program and performed the analysis; Chang Liu and Yukinori Koyama interpreted the results; Kiyoyuki Terakuta and Ryo Yoshida supervised the research.

## Additional Information

**Competing Interests:** The authors declare no competing interests.

# Supplementary Information

# Recreation of the Periodic Table with an Unsupervised Machine Learning Algorithm

**Minoru Kusaba[1,*], Chang Liu[2], Yukinori Koyama[3], Kiyoyuki Terakura[4], Ryo Yoshida[1,2,3,*]**

[1]The Graduate University for Advanced Studies, SOKENDAI, Tachikawa, Tokyo 190-8562, Japan
[2]The Institute of Statistical Mathematics, Research Organization of Information and Systems, Tachikawa, Tokyo 190-8562, Japan
[3]National Institute for Materials Science, Tsukuba, Ibaraki 305-0047, Japan
[4]National Institute of Advanced Industrial Science and Technology, Tsukuba, Ibaraki 305-8560, Japan
[*]kusaba@ism.ac.jp, yoshidar@ism.ac.jp.

Figure S1 shows a heatmap of the elements' data used in this study. Detailed description of the elements-level properties for the data is given in Figure S2. Figure S3 shows visualization results of the elements' data on the 2-dimensional space using various unsupervised learning methods. The periodic table generator (PTG) landscapes of 39 features for the square and conical tables corresponding to Figs. 2 and 3 are shown in Figures S4 and S5. From pages 9 to 12, a detailed description of GTM-LDLV is given. A summary of the algorithm of PTG is shown in Algorithm 1. Finally, details of the analysis procedure used in this study are given from pages 14 to 16.

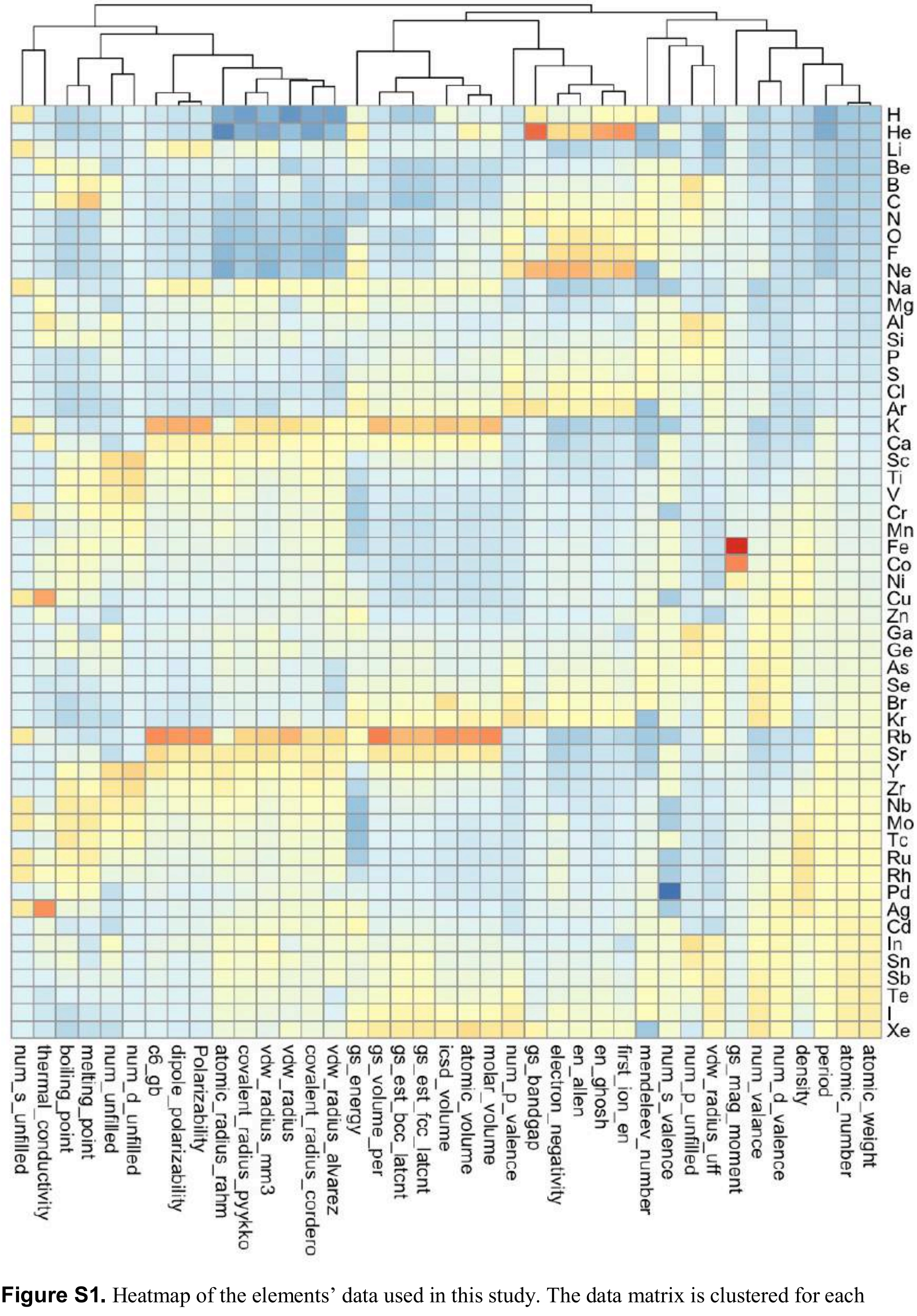

**Figure S1.** Heatmap of the elements' data used in this study. The data matrix is clustered for each column (features).

**Elements-level properties used for analysis**

| feature | description |
|---|---|
| atomic_number | Number of protons found in the nucleus of an atom |
| atomic_radius_rahm | Atomic radius by Rahm et al |
| atomic_volume | Atomic volume |
| atomic_weight | The mass of an atom |
| boiling_point | Boiling temperature |
| c6_gb | C_6 dispersion coefficient in a.u |
| covalent_radius_cordero | Covalent radius by Cerdero et al |
| covalent_radius_pyykko | Single bond covalent radius by Pyykko et al |
| density | Density at 295K |
| dipole_polarizability | Dipole polarizability |
| electron_negativity | Tendency of an atom to attract a shared pair of electrons |
| en_allen | Allen's scale of electronegativity |
| en_ghosh | Ghosh's scale of electronegativity |
| first_ion_en | First ionisation energy |
| gs_bandgap | DFT bandgap energy of T=0K ground state |
| gs_energy | DFT energy per atom (raw VASP value) of T=0K ground state |
| gs_est_bcc_latcnt | Estimated BCC lattice parameter based on the DFT volume |
| gs_est_fcc_latcnt | Estimated FCC lattice parameter based on the DFT volume |
| gs_mag_moment | DFT magnetic momenet of T=0K ground state |
| gs_volume_per | DFT volume per atom of T=0K ground state |
| icsd_volume | Atom volume in ICSD database |
| mendeleev_number | Atom number in mendeleev's periodic table |
| melting_point | Melting point |
| molar_volume | Molar volume |
| num_unfilled | Total unfilled electron |
| num_valence | Total valence electron |
| num_d_unfilled | Unfilled electron in d shell |
| num_d_valence | Valence electron in d shell |
| num_p_unfilled | Unfilled electron in p shell |
| num_p_valence | Valence electron in p shell |
| num_s_unfilled | Unfilled electron in s shell |
| num_s_valence | Valence electron in s shell |
| period | Period in the periodic table |
| thermal_conductivity | Thermal conductivity at 25 C |
| vdw_radius | Van der Waals radius |
| vdw_radius_alvarez | Van der Waals radius according to Alvarez |
| vdw_radius_mm3 | Van der Waals radius from the MM3 FF |
| vdw_radius_uff | Van der Waals radius from the UFF |
| Polarizability | Ability to form instantaneous dipoles |

**Figure S2.** Detailed description for 39 elements-level features used in this analysis.

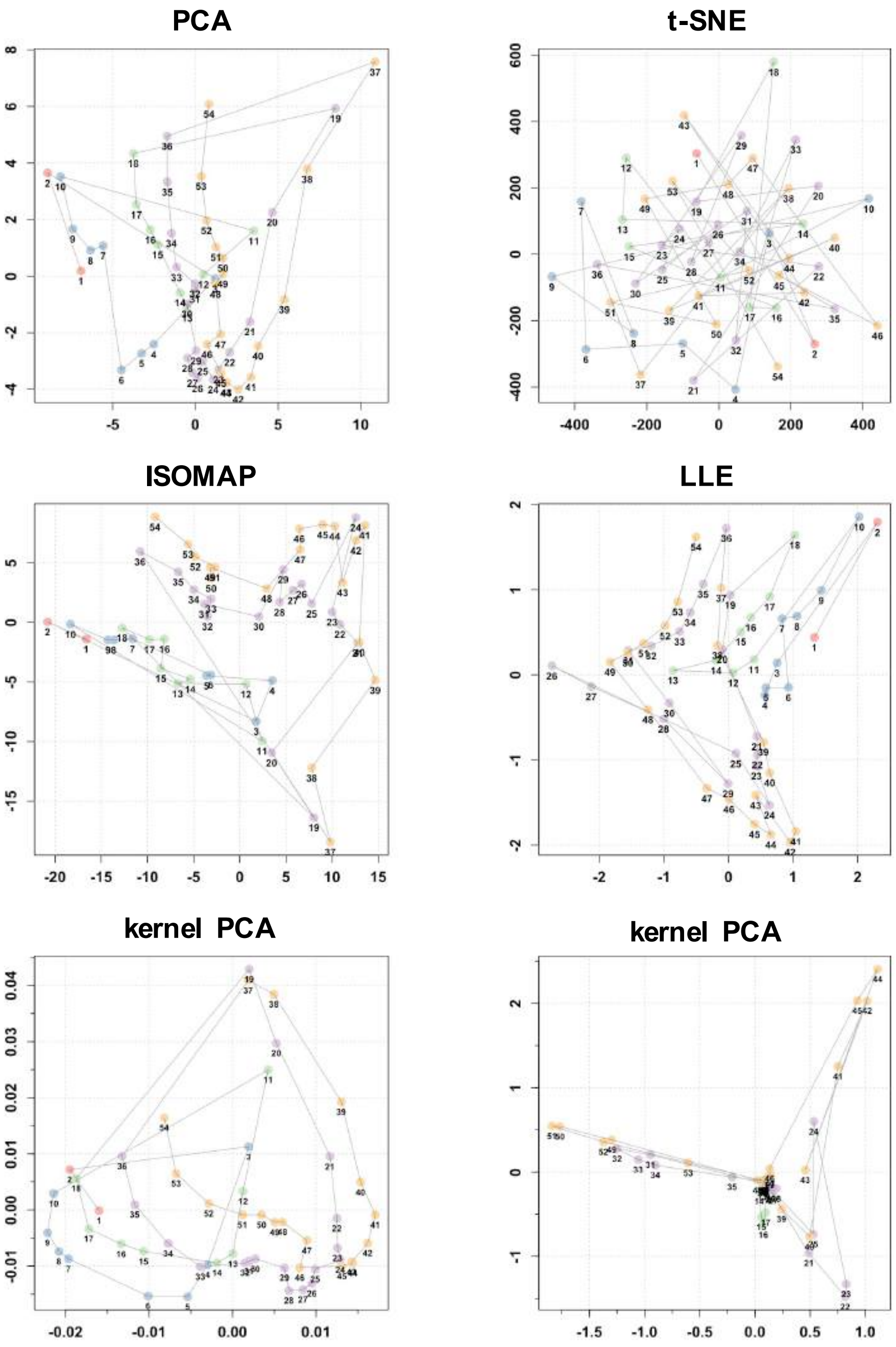

**Figure S3.** Visualization results of the elements' data on the two-dimensional space using PCA (Top-left), t-SNE (Top-right), ISOMAP with neighbors = 3 (Middle-left), LLE with neighbors = 9 (Middle-right), kernel PCA with ANOVA kernel and sigma = 0.2 (Bottom-left), and kernel PCA with RBF kernel and sigma = 0.2. The elements are colour-coded for each period and numbered by atomic number. A line passing through the elements is drawn in atomic number order.

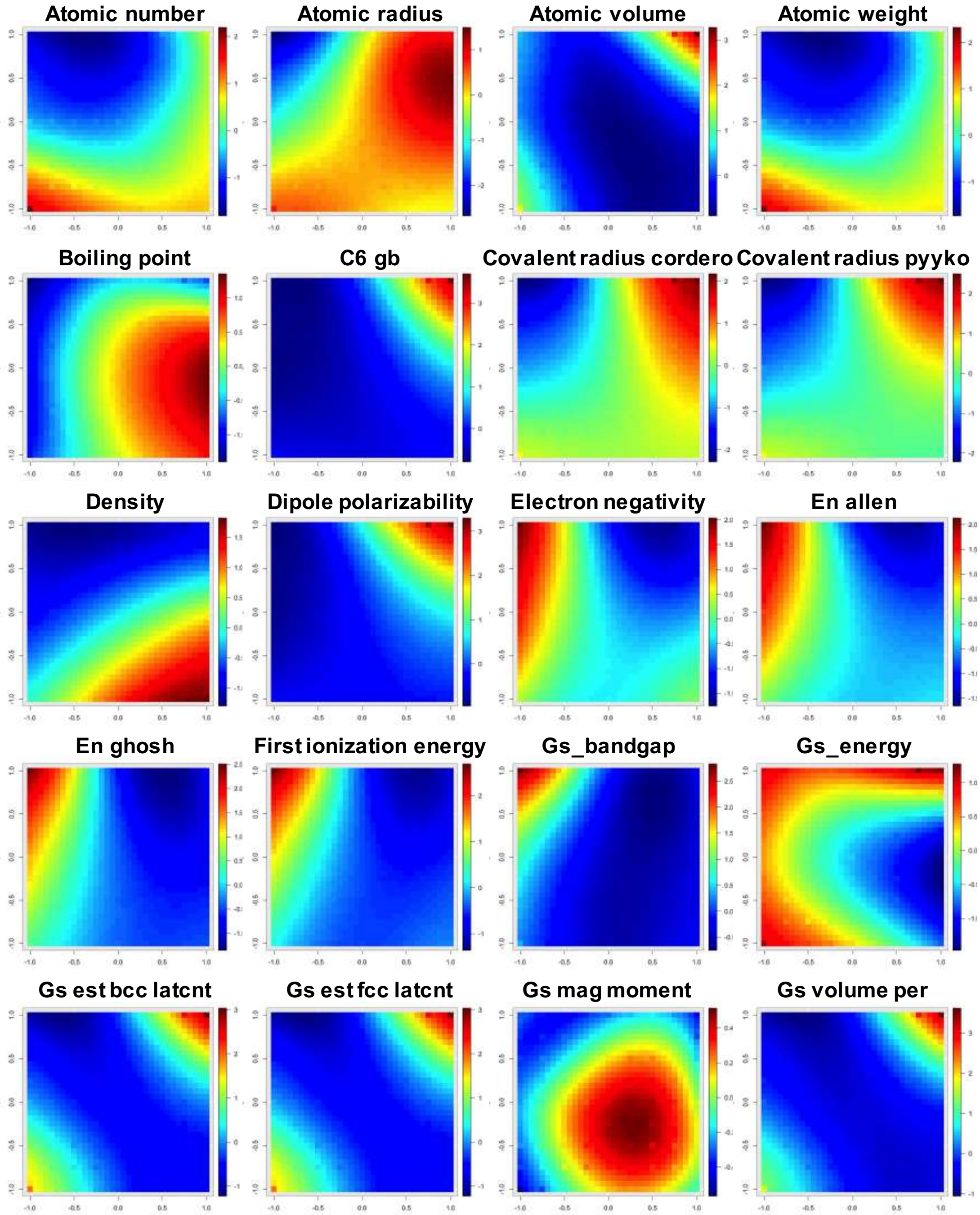

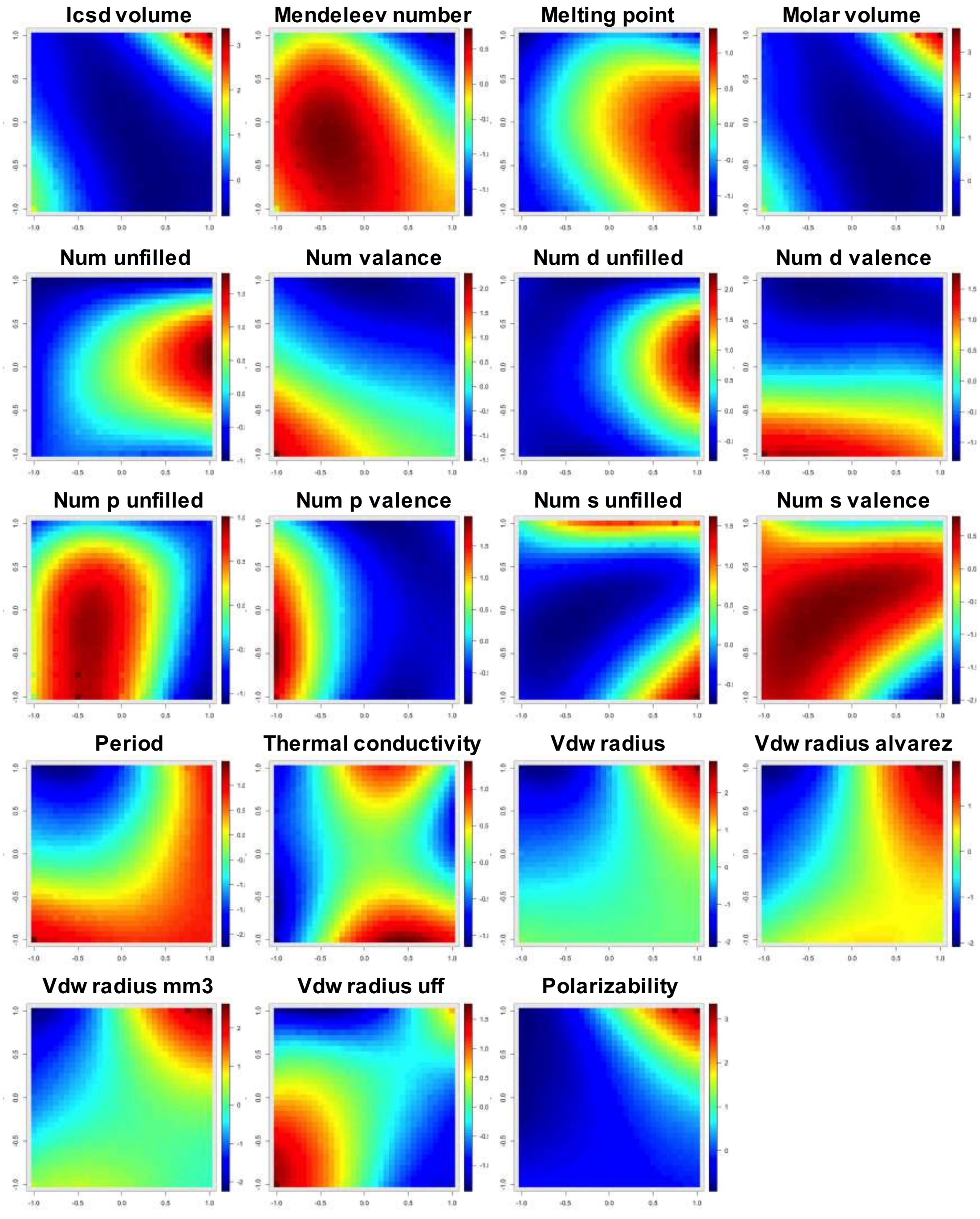

**Figure S4.** PTG property landscape of all 39 features for the square PTG table shown in Fig. 2.

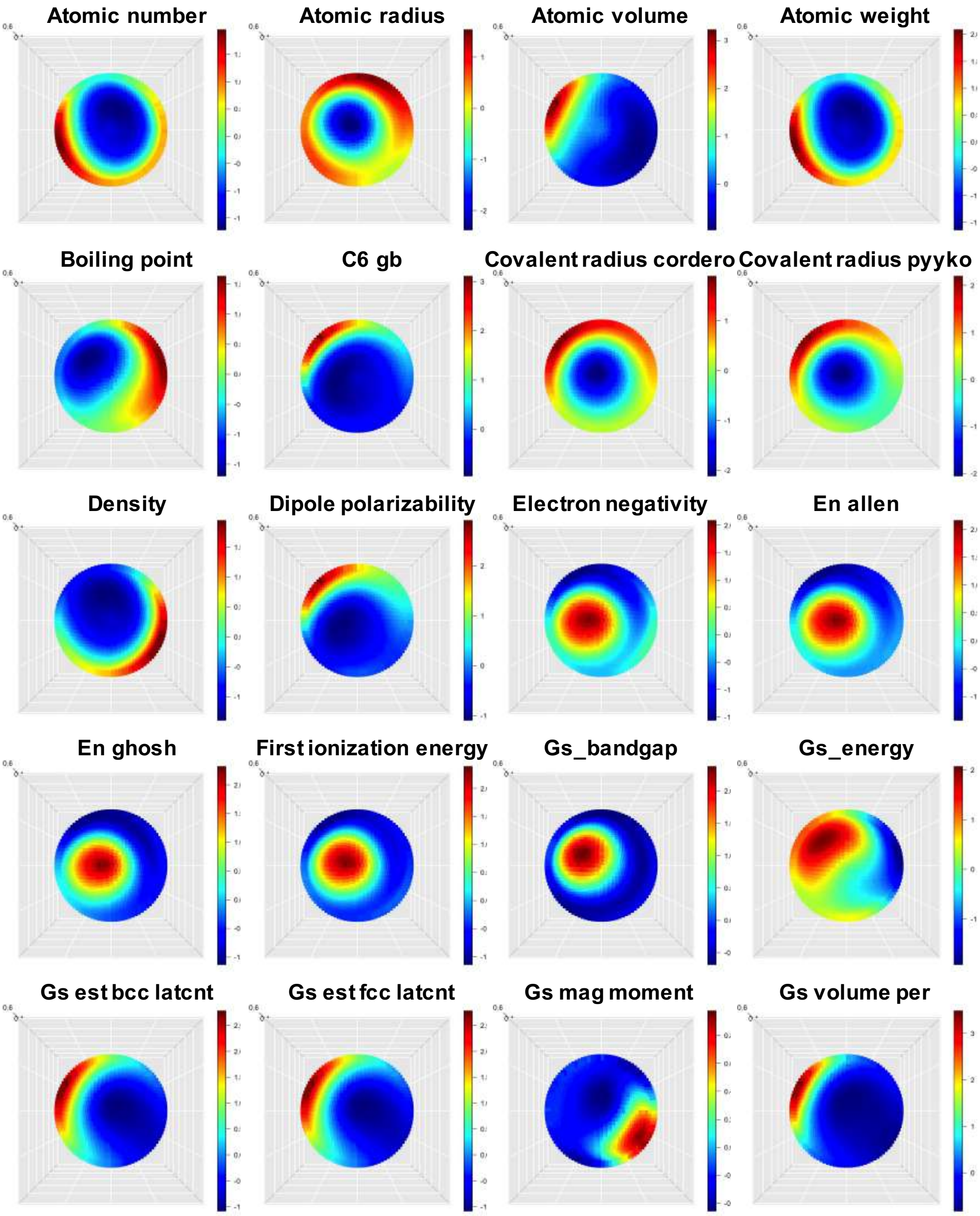

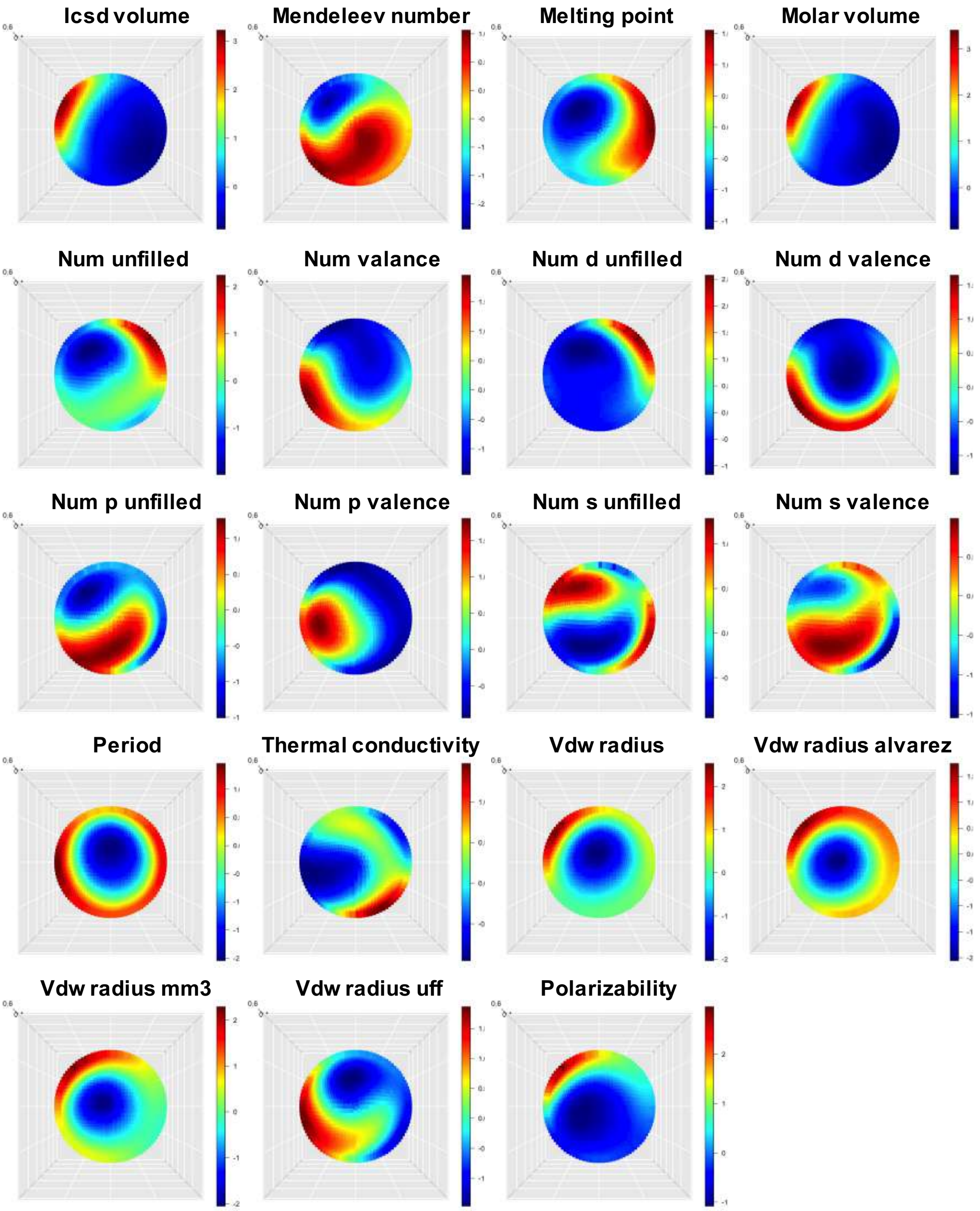

**Figure S5.** PTG property landscapes of all 39 features for the conical PTG table shown in Fig. 3.

## Detailed Method of GTM-LDLV

Our learning method can be considered as an extension of generative topographic mapping (GTM) proposed by Bishop et al. [1]. GTM is a latent variable model that represents the probability density of data using a nonlinear function of lower dimensional latent variables. It can be regarded as a stochastic formulation of self-organizing map (SOM) [2].

In GTM, $K$ grid points (called "nodes" hereafter) $\boldsymbol{u}_1, \cdots, \boldsymbol{u}_K$ regularly arranged in the $L$-dimensional latent space are prepared for data visualization, and consider a nonlinear function $\boldsymbol{f}(\boldsymbol{u}_k; \boldsymbol{\theta})$ that maps the nodes $\boldsymbol{u}_k$ to a point $\boldsymbol{y}_k$ on the $D$-dimensional feature space. The dimension of the latent space $L$ is set less than 3 for visualization. $\boldsymbol{\theta}$ is a parameter set that determines $\boldsymbol{f}(\boldsymbol{u}_k; \boldsymbol{\theta})$. It is assumed that the $D$-dimensional feature vector $\boldsymbol{x}_n$ is generated independently by a restricted mixture of $K$ Gaussian distributions, where all mixing coefficients are $1/K$, the mean of the Gaussian distribution is $\boldsymbol{y}_k$, and the covariance matrix is all $\beta^{-1}\boldsymbol{I}$. Then, the distribution is given by

$$p(\boldsymbol{x}_n|\boldsymbol{\theta}, \beta) = \frac{1}{K}\sum_{k=1}^{K} p(\boldsymbol{x}_n|\boldsymbol{u}_k, \boldsymbol{\theta}, \beta),$$

$$p(\boldsymbol{x}_n|\boldsymbol{u}_k, \boldsymbol{\theta}, \beta) = N(\boldsymbol{x}_n|\boldsymbol{y}_k, \beta^{-1}\boldsymbol{I}), \qquad \boldsymbol{y}_k = \boldsymbol{f}(\boldsymbol{u}_k; \boldsymbol{\theta}),$$

where $N(\cdot\,|\boldsymbol{\mu}, \boldsymbol{\Sigma})$ denotes the Gaussian density function with mean $\boldsymbol{\mu}$ and covariance matrix $\boldsymbol{\Sigma}$. Here, we introduce a vector of $K$ latent variables, $\boldsymbol{z}_n = (z_{1n}, \cdots, z_{Kn})'$. The $k$th entry $z_{kn}$ takes the value 1 if $\boldsymbol{x}_n$ is generated by the $k$th component distribution, and 0 otherwise. Here, let $\boldsymbol{X}$ denote a matrix of $\boldsymbol{x}_1, \cdots, \boldsymbol{x}_N$ elements, and $\boldsymbol{Z}$ be a matrix of $\boldsymbol{z}_1, \cdots, \boldsymbol{z}_N$. Then, their joint distribution is given by

$$p(\boldsymbol{X}, \boldsymbol{Z}|\boldsymbol{\theta}, \beta) = K^{-N}\prod_{n=1}^{N}\prod_{k=1}^{K} N(\boldsymbol{x}_n|\boldsymbol{y}_k, \beta^{-1}\boldsymbol{I})^{z_{Kn}}. \tag{1}$$

If the function $\boldsymbol{f}(\boldsymbol{u}_k; \boldsymbol{\theta})$ is a smooth nonlinear function, then nodes $\boldsymbol{u}_k$ are mapped onto $\boldsymbol{y}_k$ while maintaining the topological relationship in the latent space. GTM is seen as a mixture of Gaussian distributions, which means $\boldsymbol{y}_k$ are restricted to the lower dimensional manifold.

In GTM, the function $\boldsymbol{f}(\boldsymbol{u}_k; \boldsymbol{\theta})$ is constructed by a Gaussian process (GP) [3]. The nature of the GP is determined by the choice of a covariance function. The conventional GTM model uses a covariance function with a constant length scale throughout the latent space. This model cannot locally change the smoothness of the nonlinear function representing the distribution of the observed data according to the value of the latent variable. The underlying patterns of the element data are considered nonlinear and highly complex, thus we require a GTM model that can represent more flexible functions. Therefore, we focused on GTM-LDLV [4], which is a recently proposed GTM model that can control the smoothness of the nonlinear function locally according to the value of the latent variable.

In GTM-LDLV, it is assumed that the $D$-dimensional feature vector $\boldsymbol{x}_n$ is generated independently by a restricted mixture of $K$ Gaussian distributions defined in equation (1), and the nonlinear function $\boldsymbol{f}(\boldsymbol{u}_k)$ is modelled to be the product of two functions: a $D$-dimensional vector-valued function $\boldsymbol{h}(\boldsymbol{u}_k)$ and a positive scalar function $g(\boldsymbol{u}_k)$. Then, their joint distribution is given by

$$p(\boldsymbol{X},\boldsymbol{Z}|\boldsymbol{g},\boldsymbol{H},\beta) = K^{-N}\prod_{n=1}^{N}\prod_{k=1}^{K}N(\boldsymbol{x}_n|\boldsymbol{y}_k,\beta^{-1}\boldsymbol{I})^{z_{Kn}}, \qquad \boldsymbol{y}_k = \boldsymbol{f}(\boldsymbol{u}_k) = g(\boldsymbol{u}_k)\boldsymbol{h}(\boldsymbol{u}_k),$$

where $\boldsymbol{g}$ is a vector $g(\boldsymbol{u}_k)\,(k = 1,\cdots,K)$, and $\boldsymbol{H}$ is a matrix $\boldsymbol{h}(\boldsymbol{u}_k)\,(k = 1,\cdots,K)$. The prior distribution of $g(\boldsymbol{u})$ is given as a truncated GP with mean 0 and covariance function $c_g(\boldsymbol{u}_i,\boldsymbol{u}_j;\boldsymbol{\xi}_g)$, which handles positive-bounded random functions. The prior distribution of the $d$th entry $h_d(\boldsymbol{u})$ of $\boldsymbol{h}(\boldsymbol{u})$ is given as a GP with mean 0 and covariance function $c_h(\boldsymbol{u}_i,\boldsymbol{u}_j)$. The prior distribution of the parameters $g$ and $\boldsymbol{H}$ are given by

$$p(\boldsymbol{g}) = N^{+}\left(\boldsymbol{g}\middle|\boldsymbol{0},\boldsymbol{C}_g(\boldsymbol{\xi}_g)\right), \tag{2}$$

$$p(\boldsymbol{H}|r) = \prod_{d=1}^{D}N\left(\boldsymbol{h}_{(d)}|\boldsymbol{0},\boldsymbol{C}_h\right), \tag{3}$$

where $N^{+}$ is a truncated normal distribution which handles positive-bounded random functions, $\boldsymbol{h}_{(d)}$ is a vector of the $d$th entry of the matrix $\boldsymbol{H}'$, and $\boldsymbol{C}_h$ is a matrix which consists of covariance function $c_h(\boldsymbol{u}_i,\boldsymbol{u}_j)$ as an element. Specifically, the covariance functions, $c_g(\boldsymbol{u}_i,\boldsymbol{u}_j;\boldsymbol{\xi}_g)$ and $c_h(\boldsymbol{u}_i,\boldsymbol{u}_j)$, are given by

$$c_g(\boldsymbol{u}_i,\boldsymbol{u}_j;\boldsymbol{\xi}_g) = v_g \cdot \exp\left(-\frac{\|\boldsymbol{u}_i-\boldsymbol{u}_j\|^2}{2l_g}\right), \tag{4}$$

$$c_h(\boldsymbol{u}_i,\boldsymbol{u}_j) = \left\{\frac{2l(\boldsymbol{u}_i)l(\boldsymbol{u}_j)}{l^2(\boldsymbol{u}_i)+l^2(\boldsymbol{u}_j)}\right\}^{\frac{L}{2}}\exp\left(-\frac{\|\boldsymbol{u}_i-\boldsymbol{u}_j\|^2}{l^2(\boldsymbol{u}_i)+l^2(\boldsymbol{u}_j)}\right). \tag{5}$$

In equation (4), the hyperparameter $\boldsymbol{\xi}_g$ consists of $v_g$ and $l_g$, referred to as the variance and the length-scale respectively, that control the magnitude of variances and smoothness of a positive-valued function $g(\boldsymbol{u})$ generated from the GP. In equation (5), the length-scale parameter $l(\boldsymbol{u})$ is a function of $\boldsymbol{u}$ and parameterized as $l(\boldsymbol{u}) = \exp(r(\boldsymbol{u}))$ with the function $r(\boldsymbol{u})$ following GP with mean 0 and covariance function $c_r(\boldsymbol{u}_i,\boldsymbol{u}_j;\boldsymbol{\xi}_r)$. Finally, the prior distribution of the precision parameter $\beta$ is given by

$$p(\beta) = \mathrm{Gam}(\beta|d_{\beta 0},s_{\beta 0}), \tag{6}$$

where $\mathrm{Gam}(\cdot\,|d,s)$ denotes the gamma distribution, and its density function is defined by

$$\mathrm{Gam}(x|d,s) = \frac{s^d}{\Gamma(d)} x^{d-1}\exp(-sx),$$

where $\Gamma$ is the gamma function $\Gamma(x) = \int_0^\infty e^{-t}t^{x-1}\,dt$.

The unknown parameter to be estimated is $\boldsymbol{\theta} = \{\boldsymbol{Z}, \beta, \boldsymbol{g}, \boldsymbol{H}, \boldsymbol{r}\}$. In GTM-LDLV, the posterior distribution $p(\boldsymbol{\theta}|\boldsymbol{X})$ is approximately evaluated using a Markov Chain Monte Carlo (MCMC) method. Iteratively sampling from the full conditional posterior distribution for each member of $\{\boldsymbol{Z}, \beta, \boldsymbol{g}, \boldsymbol{H}, \boldsymbol{r}\}$, we obtain a set of ensembles that follow the posterior distribution approximately. By taking the ensemble average over the samples from $p(\boldsymbol{\theta}|\boldsymbol{X})$, the parameters of GTM-LDLV are estimated. The simultaneous distribution of the data $\boldsymbol{X}$ and parameters $\boldsymbol{\theta}$ is given by

$$p(\boldsymbol{X}, \boldsymbol{\theta}) = p(\boldsymbol{X}, \boldsymbol{Z}|\boldsymbol{g}, \boldsymbol{H}, \beta)p(\beta)p(\boldsymbol{g})p(\boldsymbol{H}|\boldsymbol{r})p(\boldsymbol{r}). \tag{7}$$

From equation (7) and Bayesian theorem, the posterior distribution of the latent variable $\boldsymbol{Z}$ is given by

$$p(\boldsymbol{Z}|\boldsymbol{X}, \boldsymbol{\theta}_{-\boldsymbol{Z}}) \propto p(\boldsymbol{X}, \boldsymbol{\theta}) \propto p(\boldsymbol{X}, \boldsymbol{Z}|\boldsymbol{g}, \boldsymbol{H}, \beta) \propto \prod_{n=1}^{N}\prod_{k=1}^{K} \exp\left(-\frac{\beta}{2}\|\boldsymbol{x}_n - \boldsymbol{y}_k\|^2\right)^{z_{kn}}, \tag{8}$$

where $\boldsymbol{\theta}_{-\boldsymbol{A}}$ represents a set of the parameters obtained by removing $\boldsymbol{A}$ from $\boldsymbol{\theta}$. Since summation over $k$ of $\boldsymbol{Z}$ for each $n$ is equal to 1, equation (8) can be written as

$$p(\boldsymbol{Z}|\boldsymbol{X}, \boldsymbol{\theta}_{-\boldsymbol{Z}}) = \prod_{n=1}^{N}\prod_{k=1}^{K} \gamma_k(\boldsymbol{x}_n; \boldsymbol{g}, \boldsymbol{H}, \beta)^{z_{kn}}, \tag{9}$$

where $\gamma_k(\boldsymbol{x}_n)$ is the probability that $\boldsymbol{x}_n$ is generated by the $k$th mixing element given $\boldsymbol{X}$ and $\boldsymbol{\theta}_{-\boldsymbol{Z}}$. $\gamma_k(\boldsymbol{x}_n)$ is given by

$$\gamma_k(\boldsymbol{x}_n; \boldsymbol{g}, \boldsymbol{H}, \beta) = \frac{\exp\left(-\frac{\beta}{2}\|\boldsymbol{x}_n - \boldsymbol{y}_k\|^2\right)}{\sum_{k'=1}^{K}\exp\left(-\frac{\beta}{2}\|\boldsymbol{x}_n - \boldsymbol{y}_{k'}\|^2\right)}. \tag{10}$$

Next, from equation (10) and Bayesian theorem, the conditional posterior distribution for parameters $\beta, \boldsymbol{g}, \boldsymbol{H}$, is given by

$$p(\beta|\boldsymbol{X}, \boldsymbol{\theta}_{-\beta}) = \mathrm{Gam}(\beta|d_\beta, s_\beta), \tag{11}$$

$$p(\boldsymbol{g}|\boldsymbol{X}, \boldsymbol{\theta}_{-\boldsymbol{g}}) = N^+(\boldsymbol{g}|\boldsymbol{\mu}_g, \boldsymbol{\Sigma}_g), \tag{12}$$

$$P(\boldsymbol{H}|\boldsymbol{X}, \boldsymbol{\theta}_{-\boldsymbol{H}}) = \prod_{d=1}^{D} N(\boldsymbol{h}_{(d)}|\boldsymbol{\mu}_{h,d}, \boldsymbol{\Sigma}_h). \tag{13}$$

The parameters of the conditional posterior distribution for parameters $\beta, \boldsymbol{g}, \boldsymbol{H}$ are given by

$$d_\beta = d_{\beta 0} + \frac{ND}{2},$$

$$s_\beta = s_{\beta 0} + \frac{1}{2}\sum_{n=1}^{N}\sum_{k=1}^{K} z_{kn}\|\boldsymbol{x}_n - \boldsymbol{y}_k\|^2,$$

$$\boldsymbol{\mu}_g = \beta\boldsymbol{\Sigma}_g \mathrm{diag}(\boldsymbol{Z}\boldsymbol{X}'\boldsymbol{H}),$$

$$\boldsymbol{\Sigma}_g = (\beta\boldsymbol{G}\boldsymbol{\Lambda}_h + \boldsymbol{C}_{st}(\boldsymbol{\xi}_g)^{-1})^{-1},$$

$$\boldsymbol{\mu}_{h,d} = \beta\boldsymbol{\Sigma}_h\boldsymbol{\Lambda}_g\boldsymbol{Z}\boldsymbol{x}_{(d)}, \qquad \boldsymbol{\Sigma}_h = (\beta\boldsymbol{G}\boldsymbol{\Lambda}_g^2 + \boldsymbol{C}_h^{-1})^{-1}.$$

The posterior distribution of $\boldsymbol{r}$ is given by

$$p(\boldsymbol{r}|\boldsymbol{X},\boldsymbol{\theta}_{-r}) \propto p(\boldsymbol{X},\boldsymbol{\theta}) \propto p(\boldsymbol{H}|\boldsymbol{r})p(\boldsymbol{r}) \propto \exp\big(s(\boldsymbol{r})\big),$$

$$s(\boldsymbol{r}) = -\frac{D}{2}\ln|\boldsymbol{C}_h| - \frac{1}{2}\sum_{d=1}^{D}\boldsymbol{h}'_{(d)}\boldsymbol{C}_h^{-1}\boldsymbol{h}_{(d)} - \frac{1}{2}\boldsymbol{r}'\boldsymbol{C}_{st}(\boldsymbol{\xi}_r)^{-1}\boldsymbol{r}. \qquad (14)$$

Since $\boldsymbol{C}_h$ is a matrix that depends on $\boldsymbol{r}$, a sampling of $\boldsymbol{r}$ is performed as follows using Metropolis-Hasting method [5]. First, find the local maximum point $\hat{\boldsymbol{r}}$ of the log-likelihood function $s(\boldsymbol{r})$, then generate the candidate point $\boldsymbol{r}^*$ from the proposed distribution $N(\boldsymbol{r}|\boldsymbol{m}_r,\boldsymbol{V}_r)$. $\boldsymbol{m}_r, \boldsymbol{V}_r$ are given by

$$\boldsymbol{m}_r = \hat{\boldsymbol{r}} + \boldsymbol{V}_r \left.\frac{\partial s(\boldsymbol{r})}{\partial \boldsymbol{r}}\right|_{r=\hat{r}}, \qquad \boldsymbol{V}_r = \left\{-\frac{\partial^2 s(\boldsymbol{r})}{\partial\boldsymbol{r}\partial\boldsymbol{r}'}\right\}_{r=\hat{r}}^{-1}.$$

When the current point is $\boldsymbol{r}^{t-1}$, the candidate point $\boldsymbol{r}^*$ is accepted with the next probability.

$$\min\left\{\frac{\exp\big(s(\boldsymbol{r}^*)\big)\,N(\boldsymbol{r}^{t-1}|\boldsymbol{m}_l,\boldsymbol{V}_l)}{\exp\big(s(\boldsymbol{r}^{t-1})\big)\,N(\boldsymbol{r}^*|\boldsymbol{m}_l,\boldsymbol{V}_l)}, 1\right\}. \qquad (15)$$

## Algorithm of PTG

The algorithm of PTG is summarized in Algorithm 1.

---

**Algorithm 1** Periodic Table Generator (PTG)

---

**1**: Prepare initial value $\boldsymbol{\theta}^0 = \{\boldsymbol{Z}^0, \beta^0, \boldsymbol{g}^0, \boldsymbol{H}^0, \boldsymbol{r}^0\}$.

**for** $t = 1$ to $T$ **do**

Sample $\boldsymbol{Z}^t$ from $p(\boldsymbol{Z}|\boldsymbol{X}, \beta^{t-1}, \boldsymbol{g}^{t-1}, \boldsymbol{H}^{t-1}, \boldsymbol{r}^{t-1})$.

Sample $\beta^t$ from $p(\beta|\boldsymbol{X}, \boldsymbol{Z}^t, \boldsymbol{g}^{t-1}, \boldsymbol{H}^{t-1}, \boldsymbol{r}^{t-1})$.

Sample $\boldsymbol{g}^t$ from $p(\boldsymbol{g}|\boldsymbol{X}, \boldsymbol{Z}^t, \beta^t, \boldsymbol{H}^{t-1}, \boldsymbol{r}^{t-1})$.

Sample $\boldsymbol{H}^t$ from $p(\boldsymbol{H}|\boldsymbol{X}, \boldsymbol{Z}^t, \beta^t, \boldsymbol{g}^t, \boldsymbol{r}^{t-1})$.

Sample $\boldsymbol{r}^t$ from $p(\boldsymbol{r}|\boldsymbol{X}, \boldsymbol{Z}^t, \beta^t, \boldsymbol{g}^t, \boldsymbol{H}^t)$.

**end for**

For a sufficiently large number $T_b$, record $\boldsymbol{\theta}^t = \{\boldsymbol{Z}^t, \beta^t, \boldsymbol{g}^t, \boldsymbol{H}^t, \boldsymbol{r}^t\}, t = T_b, T_b + 1, \cdots, T$.

**2**: The model parameters of GTM-LDLV $\boldsymbol{\theta}^{ldlv} = \{\boldsymbol{Z}^{ldlv}, \beta^{ldlv}, \boldsymbol{g}^{ldlv}, \boldsymbol{H}^{ldlv}, \boldsymbol{r}^{ldlv}\}$ are estimated by taking the average of $\boldsymbol{\theta}^t = \{\boldsymbol{Z}^t, \beta^t, \boldsymbol{g}^t, \boldsymbol{H}^t, \boldsymbol{r}^t\}$ for $t = T_b, T_b + 1, \cdots, T$. Increase the number of nodes on the latent space so that $K \geq N$ is satisfied. Considering the parameters estimated by GTM-LDLV (The first step of PTG) as observation values, interpolate the parameters corresponding to the newly generated nodes using GP regression.

**3**: The parameters $\boldsymbol{\theta}^{itp} = \{\boldsymbol{Z}^{itp}, \beta^{itp}, \boldsymbol{g}^{itp}, \boldsymbol{H}^{itp}, \boldsymbol{r}^{itp}\}$ obtained as above are used as initial values for the next procedure.

**for** $t = 1$ to $T'$ **do**

$\boldsymbol{Z}^t \leftarrow \underset{\boldsymbol{Z} \in A}{\operatorname{argmax}} \, p(\boldsymbol{Z}|\boldsymbol{X}, \beta^{t-1}, \boldsymbol{g}^{t-1}, \boldsymbol{H}^{t-1}, \boldsymbol{r}^{t-1}),\ A = \{\boldsymbol{Z} | \sum_{n=1}^{N} z_{kn} \leq 1\ (k = 1, \cdots, K)\}$.

$\beta^t \leftarrow \underset{\beta}{\operatorname{argmax}} \, p(\beta|\boldsymbol{X}, \boldsymbol{Z}^t, \boldsymbol{g}^{t-1}, \boldsymbol{H}^{t-1}, \boldsymbol{r}^{t-1})$.

$\boldsymbol{g}^t \leftarrow \underset{\boldsymbol{g}}{\operatorname{argmax}} \, p(\boldsymbol{g}|\boldsymbol{X}, \boldsymbol{Z}^t, \beta^t, \boldsymbol{H}^{t-1}, \boldsymbol{r}^{t-1})$.

$\boldsymbol{H}^t \leftarrow \underset{\boldsymbol{H}}{\operatorname{argmax}} \, p(\boldsymbol{H}|\boldsymbol{X}, \boldsymbol{Z}^t, \beta^t, \boldsymbol{g}^t, \boldsymbol{r}^{t-1})$.

$\boldsymbol{r}^t \leftarrow \underset{\boldsymbol{r}}{\operatorname{argmax}} \, p(\boldsymbol{r}|\boldsymbol{X}, \boldsymbol{Z}^t, \beta^t, \boldsymbol{g}^t, \boldsymbol{H}^t)$.

**end for**

---

## Notes on the PTG Algorithm

It should be noted that PTG may produce different visualization results for each trial even under the same hyper parameter settings. Indeed, PTG with the element data produced different tables for each trial of the algorithm. This implies that PTG reached around different local maxima on the likelihood surface for each trial. PTG tries to fit lower dimensional manifolds to the shape of data cloud, and there should be multiple solutions to this. Therefore, it is expected that there are many local maxima which are separated from one other on the likelihood surface of PTG. This is not counterintuitive as there should not be a unique optimal solution for arranging elements in the new periodic table. One way to deal with this problem is to run the algorithm multiple times under the same hyperparameter settings and enumerate multiple visualization results. The final result is then selected from the list of the obtained tables based on some selection criterion.

In Step 1 of PTG with the elements' data, it was observed that the learning of the model became unstable and was terminated when the non-information prior distribution was used as prior distribution of the precision $\beta$. To address the problem, a prior distribution of $\beta$ with a small scale and a sufficiently large rate was used. This prior distribution keeps the variance $\beta^{-1}$ estimated from the posterior distribution larger than a certain value, and it made the learning stable.

In the next section, we introduce details of the analysis procedure and hyper parameter settings used in this study.

## Details of Analysis Procedure

We performed PTG on two different node layouts namely, square and three-dimensional conical layouts. In the square layout of $L = 2$, we set $K = 25$ in the first step of PTG in which the $5 \times 5$ nodes were evenly arranged on the area $[-1, 1] \times [-1, 1]$. In the second step, we increased the number of nodes to $9 \times 9$ by placing new nodes at middle points on the line segments connecting between each node. In the conical layout of $L = 3,$ we first used a set of nodes with $K = 25$ that were arranged uniformly on the surface of the cone placed in the area $[-1, 1] \times [-1, 1] \times [-1, 1]$. The cone was sliced into 4 sections of the same height along the vertical axis. Then, 1 (vertex), 4, 8, and 12 (bottom) nodes were uniformly placed on the outer part of the 4 cut surfaces. In the following step, the number of slices was increased by 7, and 1 (vertex), 4, 8, 12, 16, 20, and 24 (bottom) nodes were uniformly arranged in the same way. In both cases, we set $\boldsymbol{\xi}_g = \boldsymbol{\xi}_r = (1/3,3)$, the number of iteration in MCMC was set to $T = 10{,}000$ with the burn-in step $T_b = 5{,}000$, the number of iteration in the third step of fine-tuning was set to $T = 10$, and PTG was run 10 times under the same hyper parameter settings written above.

To quantitatively evaluate the quality of the periodic tables obtained by PTG with the same hyper parameter settings and different trials, we considered using a table as an element descriptor in machine learning tasks. The modelling procedure and the data set that was used is the same as the one written in the section of 'Quantitative comparison of periodic tables'. We performed the five-fold cross-validation on the 12,373 samples for the obtained 10 periodic tables. The prediction errors for the 10 periodic tables are shown in Figure S6 for the square table and Figure S7 for the conical table. As shown in Figure S6, the 10th square periodic table gave the lowest MAE (0.533 eV/atom) out of 10 tables. Therefore, this table was chosen as the final visualization result of the square PTG table, and it corresponds to that shown in Fig. 2. Similarly, as shown in Figure S7, the 4th conical periodic table giving the lowest MAE (0.464 eV/atom) was chosen as the final visualization result of the conical PTG table, and it corresponds to that shown in Fig. 3.

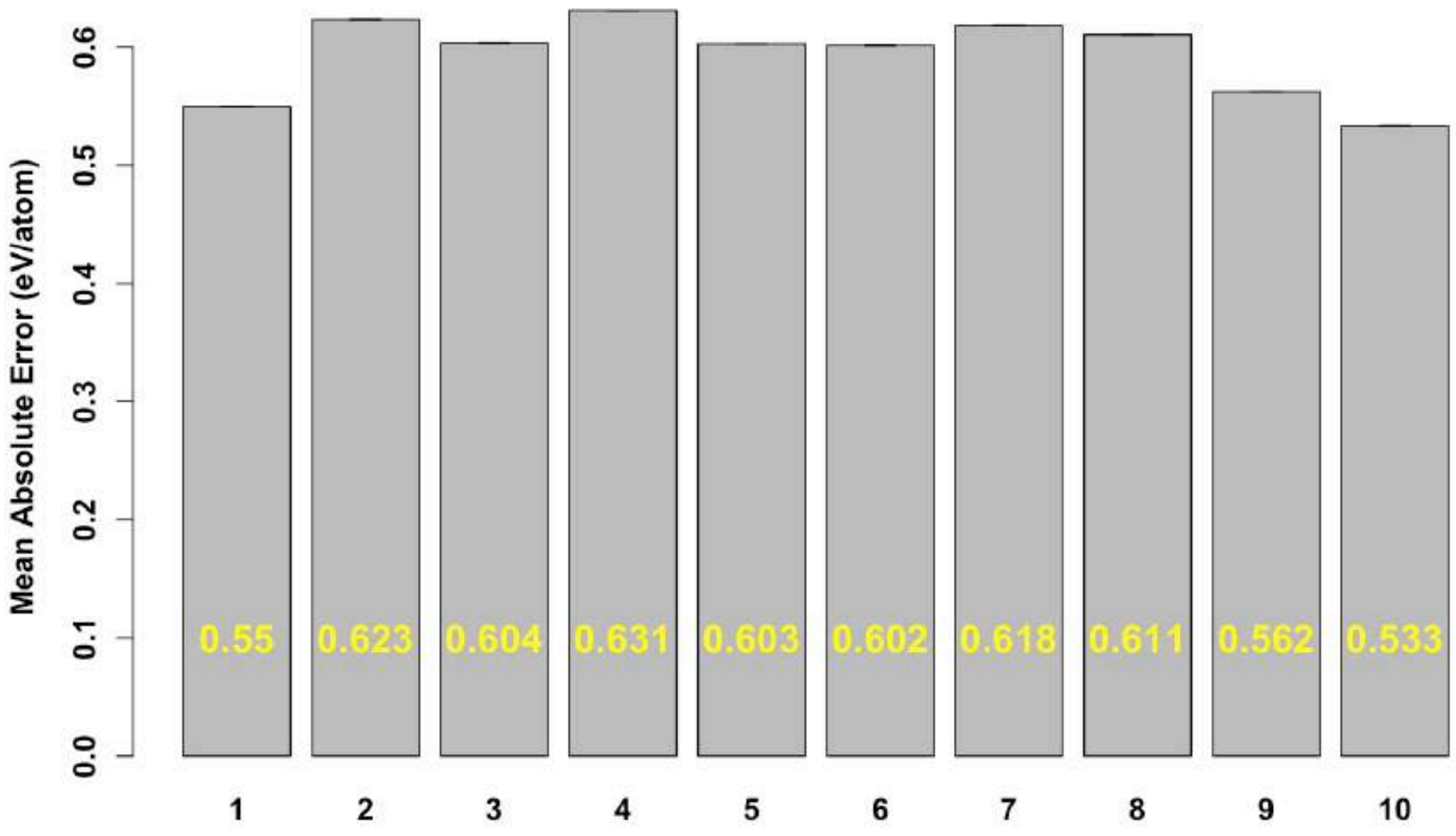

**Figure S6.** Mean absolute errors (MAE) of the prediction of the formation energy per atom for the 10 square periodic tables used as element descriptors. The vertical axis indicates cross-validated MAE of random forest regressors (RF) trained with the 10 descriptors obtained from the coordinates of elements in the square periodic tables produced by PTG, with the same hyper parameters and different trials. The error bars denote the standard deviations in 5 independent trials of the cross-validation (the error bars are invisible because of substantially small scales).

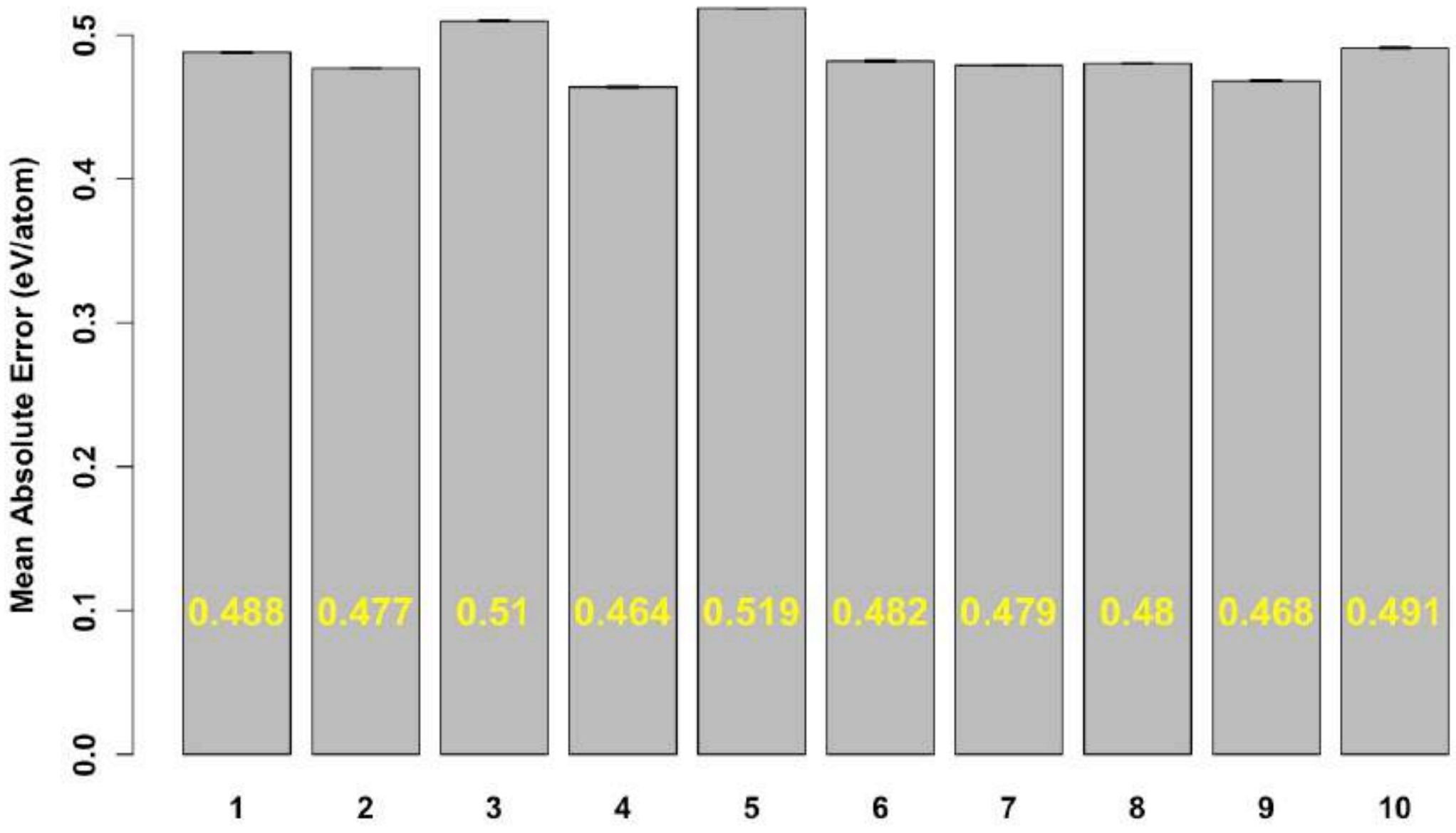

**Figure S7.** Mean absolute errors (MAE) of the prediction of the formation energy per atom for the 10 conical periodic tables used as element descriptors. The vertical axis indicates cross-validated MAE of random forest regressors (RF) trained with the 10 descriptors obtained from the coordinates of elements in the conical periodic tables produced by PTG, with the same hyper parameters and different trials. The error bars denote the standard deviations in 5 independent trials of the cross-validation (the error bars are invisible because of substantially small scales).